\documentclass[10pt,twocolumn,letterpaper]{article}

\usepackage{cvpr}              

\usepackage[english]{babel}
\usepackage{graphicx}
\usepackage{amsmath}
\usepackage{amssymb}
\usepackage{booktabs}
\usepackage{nicefrac}
\usepackage{xfrac}
\usepackage{flushend}
\usepackage{pifont}
\usepackage{balance}
\usepackage[accsupp]{axessibility}  
\newcommand{\cmark}{\ding{51}}%
\newcommand{\xmark}{\ding{55}}%

\DeclareMathOperator*{\argmin}{argmin}

\usepackage[pagebackref,breaklinks,colorlinks]{hyperref}

\usepackage[capitalize]{cleveref}
\crefname{section}{Sec.}{Secs.}
\Crefname{section}{Section}{Sections}
\Crefname{table}{Table}{Tables}
\crefname{table}{Tab.}{Tabs.}

\newcommand{\depthsfm}{D^{\text{SfM}}}
\newcommand{\bfdepthsfm}{\mathcal{D}^{\text{SfM}}}
\newcommand{\bfdepthsfmcheck}{\mathcal{D}^{\text{SfM}\checkmark}}
\newcommand{\bfdepthsfmcheckcheck}{\mathcal{D}^{\text{SfM}\checkmark \checkmark}}
\newcommand{\depthnet}{D^{\text{NN}}}
\newcommand{\bfdepthnet}{\mathbf{D}^{\text{NN}}}
\newcommand{\bfdepthnets}{\mathcal{D}^{\text{NNs}}}
\newcommand{\bfdepthnetscheck}{\mathcal{D}^{\text{NNs}\checkmark}}
\newcommand{\bfdepthnetscheckcheck}{\mathcal{D}^{\text{NNs}\checkmark \checkmark}}

\newcommand{\expresult}{~$\ast$}
\newcommand{\paperresult}{~$\diamond$}

\def\cvprPaperID{8040} 
\def\confName{CVPR}
\def\confYear{2023}

\begin{document}

\title{SfM-TTR: Using Structure from Motion for Test-Time Refinement of Single-View Depth Networks}

\author{Sergio Izquierdo \qquad Javier Civera\\
I3A, University of Zaragoza, Spain\\
{\tt\small \{izquierdo, jcivera\}@unizar.es}
}
\maketitle

\begin{abstract}
Estimating a dense depth map from a single view is geometrically ill-posed, and state-of-the-art methods rely on learning depth's relation with visual appearance using deep neural networks. On the other hand, Structure from Motion (SfM) leverages multi-view constraints to produce very accurate but sparse maps, as matching across images is typically limited by locally discriminative texture.
In this work, we combine the strengths of both approaches by proposing a novel test-time refinement (TTR) method, denoted as SfM-TTR, that boosts the performance of single-view depth networks at test time using SfM multi-view cues. Specifically, and differently from the state of the art, we use sparse SfM point clouds as test-time self-supervisory signal, fine-tuning the network encoder to learn a better representation of the test scene. Our results show how the addition of SfM-TTR to several state-of-the-art self-supervised and supervised networks improves significantly their performance, outperforming previous TTR baselines mainly based on photometric multi-view consistency. The code is available at \url{https://github.com/serizba/SfM-TTR}.

\end{abstract}

\section{Introduction}
\label{sec:intro}
Obtaining accurate and dense depth maps from images is a challenging research problem and an essential input in a wide array of fields, like robotics~\cite{zhu2022autonomous}, augmented reality~\cite{luo2020consistent}, endoscopy~\cite{recasens2021endo}, or autonomous driving~\cite{guizilini2022full}. Single-view per-pixel depth estimation is even more challenging, as it is geometrically ill-posed in the general case. However, in the last decade, intense research on deep models applied to this task has produced impressive results, showing high promise for real-world applications.

\begin{figure}[!t]
\centering
 \centering
 \includegraphics[width=\linewidth]{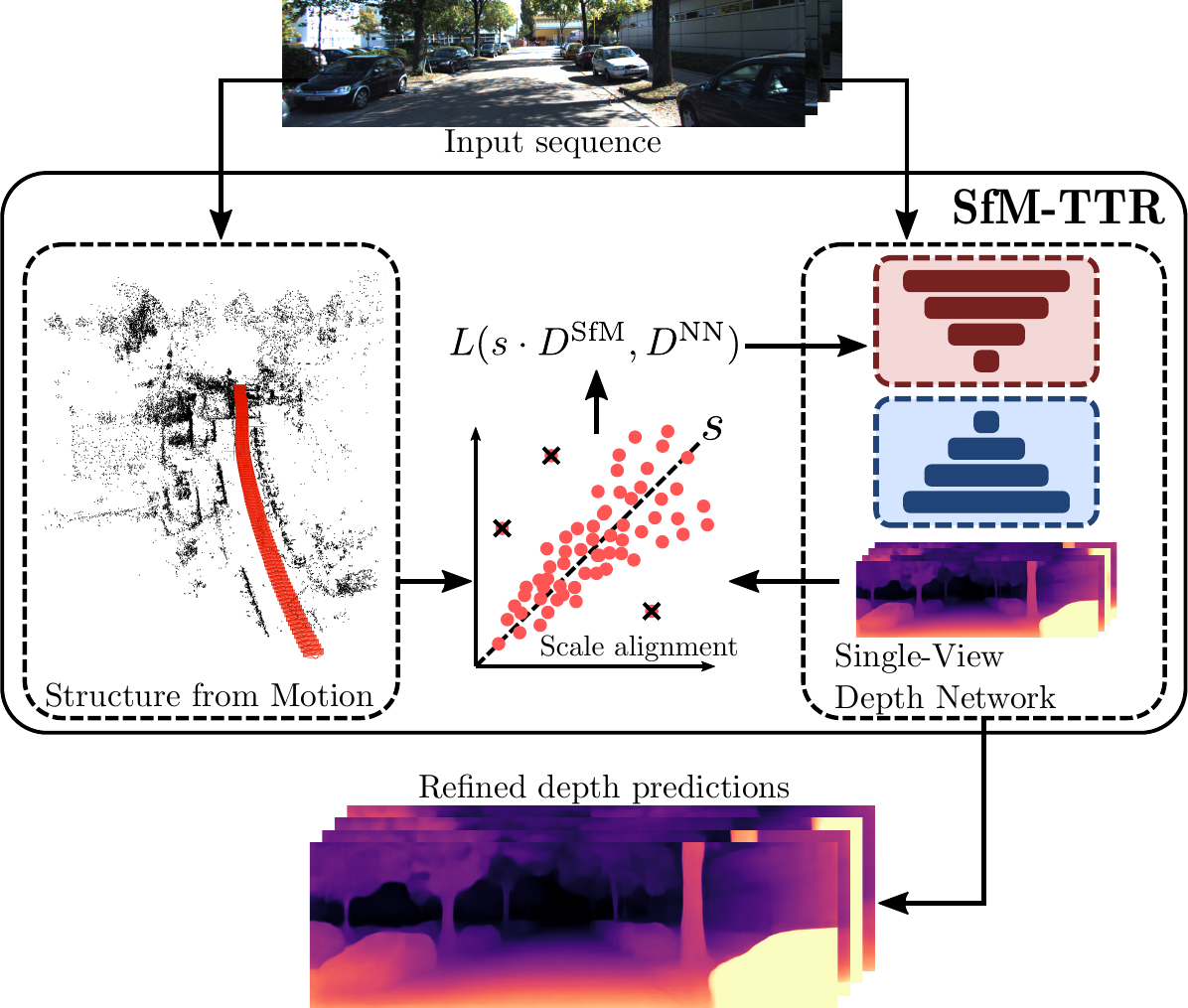}
 \vspace{.3em}
 \caption{\textbf{SfM-TTR overview}. Our approach assumes an existing pre-trained depth network and an input sequence at test time. We estimate a SfM 3D reconstruction using the input sequence, and depth maps using a single-view depth network. We align the SfM point cloud with the network's depth to obtain a pseudo-ground truth to refine the network encoder, improving its representation of the test scene and producing significantly more accurate depth estimates.}
 \label{fig:teaser}
\end{figure}

Single-view depth learning was initially addressed as a supervised learning problem, in which deep networks were trained using large image collections annotated with ground truth depth from range (e.g., LiDAR) sensors~\cite{eigen2015predicting,laina2016deeper}. At present, this line of research keeps improving the accuracy of single-view depth estimates by better learning models and training methods, as illustrated for example by~\cite{bhat2021adabins,yuan2022neural}.

In parallel to improving the learning side of the problem, several works are incorporating single- and multi-view geometric concepts to depth learning, extending its reach to more general setups. For example, \cite{facil2019cam,antequera2020mapillary} propose camera intrinsics-aware models, enabling learning and predicting depths for very different cameras. More importantly, many other works (e.g.~\cite{godard2019digging}) use losses based on multi-view photometric consistency, enabling self-supervised learning of depth and even camera intrinsics~\cite{gordon2019depth}.

Incorporating single- and multi-view geometry into depth learning naturally links the field to classic research on Structure from Motion (SfM)~\cite{Hartley2004,schonberger2016structure}, visual odometry~\cite{scaramuzza2011visual,engel2017direct} and visual SLAM~\cite{cadena2016past,campos2021orb}. These methods typically produce very accurate but sparse or semi-dense reconstructions of high-gradient points using only multi-view geometry at test time. Among the many opportunities for cross-fertilization of both fields (e.g., using depth networks in visual SLAM~\cite{tateno2017cnn} or SfM for training depth networks~\cite{li2018megadepth, klodt2018supervising, yang2018deep}), our work focuses on using SfM for refining single-view depth networks at test time.

As single-view depth applications typically include a moving camera, several recent works incorporate multiple views at inference or refine single-view depth networks with multi-view consistency cues~\cite{chen2019self,mccraith2020monocular,luo2020consistent,shu2020feature,tiwari2020pseudo,watson2021temporal,bae2022multi}.
Most approaches, however, rely mainly on photometric losses, similar to the ones used for self-supervised training. These losses are limited to be computed between close views, creating weak geometric constraints. Our contribution in this paper is a novel method that, differently from the others in the literature, uses exclusively a SfM reconstruction for TTR. Although SfM supervision is sparser than typical photometric losses, it is also significantly less noisy as it has been estimated from wider baselines. Our results show that our approach, which we denote as SfM-TTR, provides state-of-the-art results for TTR, outperforming photometric test-time refinement (Ph-TTR) for several state-of the-art supervised and self-supervised baselines.

\section{Related Work}
\label{sec:related}

Although there exists a large corpus of work on single-view depth under certain assumptions on the scene geometry, e.g.~\cite{sturm1999method,hoiem2005geometric,barinova2008fast,pan2015inferring,zaheer2018single,ali2020real}, and on multi-view depth, e.g.~\cite{zhang2009consistent, karsch2014depth}, we focus here on approaches that are mainly based on learning and target general scenes.

\subsection{Supervised Single-View Depth Learning}
Several early works addressed single-view depth learning either directly from the image~\cite{saxena2008make3d} or via semantic labels~\cite{liu2010single} before the deep learning era. The seminal works by Eigen et al.~\cite{eigen2014depth,eigen2015predicting} significantly improved the prediction accuracy by training deep networks supervised with ground-truth depth from range sensors. Since then, single-view depth networks have received significant attention from the research community, focusing on improving the performance by using more sophisticated architectures and losses, e.g.,~\cite{laina2016deeper,liu2015learning,xu2018structured,lee2018single,xian2020structure,miangoleh2021boosting,patil2022p3depth}. A re-formulation of the problem as an ordinal regression has led to further improvement~\cite{fu2018deep,bhat2021adabins,bhat2022localbins}. Recently, Bae et al.\cite{bae2022multi} fuse the single-view depths from multiple images, but differently from us without TTR of the network.
Despite their remarkable progress, we show in this paper that the accuracy of state-of-the-art supervised depth networks is further improved by our SfM-TTR proposal.

\subsection{Self-Supervised Single-View Depth Learning}
\label{sec:relatedself}

As ground truth depth annotations are uncommon, self-supervised approaches emerged as an alternative, exploiting multi-view photometric consistency~\cite{zhou2017unsupervised, godard2017unsupervised}.
Attracted by the convenience of training without depth labels, many works have further focused on addressing this paradigm, e.g.,~\cite{yin2018geonet,yang2018unsupervised,zhan2018unsupervised,liu2019neural,johnston2020self,shu2020feature,zhou2022self}. Close to our work, SfM has been used as supervisory signal during training, but limited to probabilistic networks \cite{klodt2018supervising}, or using disparities~\cite{yang2018deep} that require stereo images.
Among self-supervised works, Monodepth2~\cite{godard2019digging}, which proposed a robust loss to handle occlusions and discard invalid pixels, is of particular relevance. Monodepth2 is the base of most state-of-the-art approaches, and specifically of the baselines we chose to validate SfM-TTR: CADepth~\cite{yan2021channel}, that uses self-attention to capture more context, DIFFNet~\cite{zhou2021diffnet}, that applies feature fusion to incorporate semantic information, and ManyDepth~\cite{watson2021temporal}, that leverages more than one frame at inference to improve the predictions. All these networks use a typical encoder-decoder architecture and can be seamlessly refined with our SfM-TTR method.

\subsection{Test-Time Refinement (TTR)}
Multi-view consistency is the basis for both self-supervised depth learning and bundle adjustment~\cite{triggs1999bundle}, this last one naturally occurring at test time. 
Inspired by that, TTR was proposed~\cite{chen2019self, casser2019depth}, updating the network with the same self-supervised losses from training.
Similarly, McCraith et al.~\cite{mccraith2020monocular} showed the benefits of encoder-only fine-tuning and proposed two TTR modes: sequence- and instance-wise. Similar approaches were presented by Watson et al.~\cite{watson2021temporal}, with multiple input images for the network, Shu et al.~\cite{shu2020feature}, with a feature-metric loss, and Kuznietsov et al.~\cite{kuznietsov2021comoda}, using a replay buffer. All these TTR methods inherit the small baseline limitations from photometric losses, showing small improvements for medium and large depths for which close views produce small parallax. 
At these depths, our SfM-TTR introduces wide baseline cues, due to the higher invariance of features matching at wide baselines.
This leads to significant improvements over the state of the art.

Tiwari et al.~\cite{tiwari2020pseudo} iterates over optimizing the parameters of a single-view depth network and running pseudo-RGBD SLAM for pose estimation, but their alignment ignores the depth distributions, which results in smaller improvements compared to ours.
Luo et al. work ~\cite{luo2020consistent} is more related to ours, using SfM and optical flow as geometric constraints. 
However, despite heavy optimization (taking up to 40 minutes for a sequence of less than $250$ frames), their TTR cannot improve over baseline networks on KITTI. Instead of defining derived constraints, we directly optimize the encoder using the sparse reconstruction as pseudo ground truth, resulting in a lighter and more effective pipeline.

\section{SfM-TTR}
\label{sec:method}

Our SfM-TTR takes \emph{any} single-view depth network, trained either supervised or self-supervisedly, and fine-tunes it for the test data by a three-stage process. As a brief summary, we first estimate a sparse feature-based reconstruction of the scene from multiple views (Section \ref{sec:method_obtaining}) and predict depth outputs with the network (Section \ref{sec:method_obtaining_depthnn}). Then, we align the scale of the sparse point cloud and the network's depth (Section \ref{sec:scaling}). Finally, we fine-tune the network using the depths of the aligned sparse point cloud as supervisory signal (Section \ref{sec:sfmttr}).

\subsection{Multi-View Depth from SfM}
\label{sec:method_obtaining}

We perform a 3D reconstruction of the target scene using an off-the-shelf SfM algorithm. In our current implementation we use COLMAP~\cite{schonberger2016structure}, as it shows a high degree of accuracy and robustness in a wide variety of scenarios, although alternative SfM or visual SLAM implementations could also have been used~\cite{moulon2016openmvg,OpenSfM,campos2021orb}.

From a set of images $\mathcal{I}=\{\mathbf{I}_1, \ldots, \mathbf{I}_K\}$, $\mathbf{I}_k \in \mathbb{R}^{w \times h \times 3} \ \forall k \in \{1, \ldots, K\}$ of a scene, COLMAP returns a set of $J$ 6-degrees-of-freedom poses $\mathcal{P} = \{\mathbf{P}_1, \ldots, \mathbf{P}_J\}$, $\mathbf{P}_j=\big(\begin{smallmatrix}
  \mathbf{R}_j & \mathbf{t}_j\\
  \mathbf{0} & 1
\end{smallmatrix}\big) \in \mathbf{SE}(3) \ \forall j \in \{1, \ldots, J\}$, $J \le K$, corresponding to the cameras that the method was able to register, and the set of 3D keypoints $\mathcal{X} = \{\mathbf{X}_1, \ldots, \mathbf{X}_I\}$, $\mathbf{X}_i \in \mathbb{R}^3 \ \forall i \in \{1, \ldots, I\}$ that were reconstructed, all of them in a common reference frame. The camera with pose $\mathbf{P}_j$ observes a subset of $L_j$ points from the total set of 3D points $\mathcal{X}_j = \{\mathbf{X}_1, \ldots, \mathbf{X}_{L_j}\}\subset\mathcal{X}$. COLMAP final estimates are obtained by minimizing the sum of the squared reprojection errors $\sum_{j=1}^J \sum_{l=1}^{L_j} \mathbf{r}_{l,j}^2$.

The depth of each of the $l^\text{th}$ point in the $j^\text{th}$ camera frame is computed as 

\begin{equation}
    \depthsfm_{l,j} = \mathbf{e}_3^\top \left( \mathbf{R}_j^\top \left( \mathbf{X}_l - \mathbf{t}_j \right) \right)
\end{equation}

\noindent where $\mathbf{e}_3 = \begin{pmatrix} 0 & 0 & 1 \end{pmatrix}^\top$ is the unit vector in the optical axis direction. We will group the depths for the sparse set of points $\mathcal{X}_j$ in the set $\bfdepthsfm_j = \{ \depthsfm_{1,j}, \ldots, \depthsfm_{L_j,j}\}$, $\depthsfm_{l,j} \in \mathbb{R}_{>0} \ \forall l \in \{1, \ldots, L_j\}$, and the depths for all images in $\bfdepthsfm = \{ \bfdepthsfm_{1}, \ldots, \bfdepthsfm_{J}\} \ \forall j \in \{1, \ldots, J\}$.

\subsection{Single-View Depth from Neural Networks}
\label{sec:method_obtaining_depthnn}

Our SfM-TTR method can be applied to any architecture, and hence its predicted depth $\bfdepthnet_j \in \mathbb{R}^{w \times h}$ for an image $\mathbf{I}_j$ can be generally formulated as 

\begin{equation}
    \bfdepthnet_j = h\left( g \left( \mathbf{I}_j, \boldsymbol{\theta}_g \right), \boldsymbol{\theta}_h \right)
\end{equation}

\noindent where $h(\cdot)$ and $g(\cdot)$ stand respectively for the decoder and encoder parts of the deep networks, and $\boldsymbol{\theta}_h$ and $\boldsymbol{\theta}_g$ their respective weights, that have been trained either supervised or self-supervisedly. 

Note that the depths $\bfdepthsfm_j$ and $\bfdepthnet_j$ correspond to the same image $\mathbf{I}_j$ but are respectively sparse and dense, having hence a different number of elements, and they may have different scales. The scale is unobservable by COLMAP and self-supervised networks, while it is learned from the training data by supervised networks.

In order to estimate the relative scale between $\bfdepthsfm_j$ and $\bfdepthnet_j$ and refine at inference time the deep network, we have to select from $\bfdepthnet_j$ those elements corresponding to the sparse depth of $\bfdepthsfm_j$. For a general element $l$, we use the sampling operator $\left[ \cdot \right]$ to access the depth corresponding to the pixel coordinates $\mathbf{p}_{l,j}$

\begin{equation}
    \depthnet_{l,j} = \bfdepthnet_j \left[ \mathbf{p}_{l,j} \right]
\end{equation}

\noindent where $\mathbf{p}_{l,j}$ is obtained from the coordinates of the 3D points $\mathbf{X}_l \in \mathcal{X}_j$ and the camera pose $\mathbf{P}_j$ and applying the pinhole projection function, that we will denote as $\pi(\cdot)$

\begin{equation}
    \mathbf{p}_{l,j} = \begin{pmatrix} u & v \end{pmatrix}^\top_{l,j} = \pi \left( \mathbf{R}_j^\top \left( \mathbf{X}_l - \mathbf{t}_j \right) \right)
\end{equation}

We finally group the depths predicted by the deep network for the sparse set of points $\mathcal{X}_j$ in a joint set $\bfdepthnets_j = \{ \depthnet_{1,j}, \ldots, \depthnet_{L_j,j} \}$, $\depthnet_{l,j} \in \mathbb{R}_{>0}$ , and the depths for all images in $\bfdepthnets = \{ \bfdepthnets_{1}, \ldots, \bfdepthnets_{J}\} \ \forall j \in \{1, \ldots, J\}$.

\subsection{Scale Alignment}
\label{sec:scaling}

Scale alignment is not trivial in our setup, as both $\bfdepthsfm$ and $\bfdepthnets$ are affected by heteroscedastic (depth-dependent) inlier noise and contain a non-negligible rate of outliers. In addition, we are interested in removing outliers from $\bfdepthsfm$, but we do want to keep them in $\bfdepthnets$, as then our SfM-TTR can reduce their errors. We developed a novel scale alignment method with two stages: we make a first fit with a strict inlier model to obtain an accurate relative scale, and then relax it in the second stage to select the points used for self-supervision from $\bfdepthsfm$.

In the first stage we use RANSAC~\cite{fischler1981random}, computing 1D model instantiations, $s_{l,j}=\nicefrac{\depthnet_{l,j}}{\depthsfm_{l,j}}$ and consider in the inlier set $\bfdepthnetscheck \subset \bfdepthnets$ and $\bfdepthsfmcheck \subset \bfdepthsfm$ all depths pairs $\{\depthnet_{l^\prime,j^\prime},\depthsfm_{l^\prime,j^\prime}\}$ for which the following holds

\begin{equation}
\label{eq:scaleransac}
    \frac{\left( s_{l,j} \cdot \depthsfm_{l^\prime,j^\prime} - \depthnet_{l^\prime,j^\prime} \right)^2}{s_{l,j} \cdot \depthsfm_{l^\prime,j^\prime}} \leq \tau
\end{equation}

\noindent where $\tau$ is the inlier threshold.

In most occasions, the distribution of depths in the image is highly unbalanced, with higher frequencies for closer depths. This, together with the heteroscedasticity of the depth errors (errors are smaller for closer depths), causes that the frequently used median scale~\cite{luo2020consistent} corresponds to close points, biasing the estimation. Using least squares with all the inlier set $\{ \bfdepthnetscheck, \bfdepthsfmcheck \}$ is not a good alternative either, the fit will be biased in this case towards large depths as they have larger errors. For these reasons, we use weighted least squares to obtain a refined estimate of $s$ with the depths $D_{l,j}^{\text{NN}\checkmark} \in \bfdepthnetscheck$ and $D_{l,j}^{\text{SfM}\checkmark} \in \bfdepthsfmcheck$

\vspace{-1mm}
\begin{equation}
\label{eq:sjfirst}
\hat{s} = \argmin_{s} \sum_{j}\sum _{l} w_{l,j}^s\left( s \cdot D_{l,j}^{\text{SfM}\checkmark} - D_{l,j}^{\text{NN}\checkmark} \right)^{2}
\end{equation}

\noindent where $w_{l,j}^s$ is a per-pixel weight, that should be proportional to the inverse of the expected depth variance $\sigma_{l,j}^2$. Under the reasonable assumption of similar baselines and matching noises for all reconstructed points, it is well known that the variance grows with the depth squared~\cite{Hartley2004} and hence we can use as weights

\begin{equation}
w_{l,j}^s = \sfrac{1}{\sigma_{l,j}^2} \approx \sfrac{1}{\left(D_{l,j}^{\text{NN}\checkmark}\right)^2}
\end{equation}

Finally, we use $s_j$ from the optimization in Equation \ref{eq:sjfirst} to obtain the final set of inliers $\{ \bfdepthnetscheckcheck, \bfdepthsfmcheckcheck \}$ that we will use for our SfM-TTR. We proceed similarly to Equation \ref{eq:scaleransac}, but this time using the absolute value in the numerator, relaxing in this manner the model and favoring the inclusion of noisy depth predictions from the network depth set $\bfdepthnets$ in order to have the chance to improve them at test time.

\subsection{Test-Time Refinement}
\label{sec:sfmttr}

We refine the target network for the selected scene by updating its parameters using the depths in the final inlier set $\bfdepthsfmcheckcheck$ as supervision. As in \cite{luo2020consistent}, we optimize over the complete scene, thus obtaining a refined network with more consistent predictions across all views. This is different from other TTR works, such as~\cite{mccraith2020monocular}, in which they refine a different network for each frame of the sequence.

Each batch update works as follows. We sample an image $\mathbf{I}_j$ from the sequence and do a feed-forward pass through the network to obtain the depth prediction $\bfdepthnet_j$. Then we supervise the prediction with the sparse pseudo ground truth $\bfdepthsfmcheckcheck_j$. This supervision is weighted according to the reliability of the reconstructed 3D points, that we approximate based on their reprojection errors as $w_{l,j}^{\boldsymbol{\theta}}=\text{exp}(-\|\mathbf{r}_{l,j}\|_2^2)$. 

\begin{equation}
    \mathcal{L} = \frac{1}{|\bfdepthsfmcheckcheck_j|}\sum_{l}{w_{l,j}^{\boldsymbol{\theta}}\|\hat{s} \cdot D_{l,j}^{\text{SfM}\checkmark\checkmark} - D_{l,j}^{\text{NN}\checkmark\checkmark} \|_1 }
\end{equation}

As state-of-the-art depth networks already produce sharp predictions with well-defined object contours, we argue that our refinement should only optimize the internal understanding of the scene. Hence, we follow a similar approach as \cite{mccraith2020monocular} and only update the encoder parameters during the TTR, keeping the rest of the network fixed. Our TTR optimization can be hence formulated as $\hat{\boldsymbol{\theta}}_g = \argmin_{\boldsymbol{\theta}_g} \mathcal{L}$. In this manner, the frozen decoder $h(\cdot)$ keeps producing sharp predictions, but now they stem from a more informed representation of the underlying scene.

\section{Experimental Results}
\label{sec:experiments}

\subsection{Implementation Details and Baselines}

We validate our proposed SfM-TTR by applying it to different state-of-the-art baselines.
Specifically, we provide evaluations with the baselines CADepth~\cite{yan2021channel}, DIFFNet~\cite{zhou2021diffnet}, and ManyDepth~\cite{watson2021temporal} as representative of self-supervised approaches. We also implemented it on \mbox{AdaBins}~\cite{bhat2021adabins} to benchmark SfM-TTR's performance also with a representative supervised model. The same set of hyperparameters was used for SfM-TTR with all baselines, achieving a substantial improvement in all of them without requiring individual tuning.

For the sparse reconstruction, we run COLMAP~\cite{schonberger2016structure} with its default parameters, using a single pinhole camera model per sequence and sequential matching. Although we use all available images from a sequence to create the sparse reconstruction, the network is only optimized with the target frames of the evaluation. Regarding our scale alignment, we detect outliers running RANSAC for $20$ iterations with inlier threshold $\tau=0.5$. For the TTR optimization, we use Adam~\cite{kingma2014adam} applied to the encoder parameters, $\boldsymbol{\theta}_g$, with a learning rate of $10^{-4}$ for $200$ steps.

For comparison, we also implemented the instance-wise photometric refinement (Ph-TTR) from ManyDepth~\cite{watson2021temporal}\footnote{The TTR code was not available in the authors' repository at the time of submission.}, based on the work of McCraith et al.~\cite{mccraith2020monocular}, which updates the weights of the network encoder during inference using the photometric loss from the training. Table~\ref{tab:ph_ttr_impl} validates our implementation, showing similar performance as the one reported by the authors in~\cite{watson2021temporal}.

\begin{table}[h]
  \centering
  \footnotesize
  \resizebox{\linewidth}{!}{
    \begin{tabular}{|l|c|c|c|c|c|c|c|}\hline
        Method & Abs Rel~$\downarrow$ & Sq Rel~$\downarrow$ & RMSE~$\downarrow$ & RMSE log~$\downarrow$\\
    \hline\hline
ManyDepth~\cite{watson2021temporal} Ph-TTR\paperresult & 0.087  &  0.696  &  4.183  & 0.167  \\
ManyDepth~\cite{watson2021temporal} Ph-TTR\expresult & 0.088  & 0.681 & 4.122 & 0.168   \\
    \hline
    \end{tabular}
    }
  \caption{ManyDepth Ph-TTR~\cite{watson2021temporal} ($\diamond$) and our own implementation ($\ast$) obtain similar metrics.} 
  \label{tab:ph_ttr_impl}
\end{table}

\subsection{Dataset}

We run all evaluations on the KITTI dataset~\cite{geiger2012we}, the common benchmark for single- and multi-view depth learning. Regarding the KITTI ground truth for depth learning evaluations, the literature is split among those following Eigen et al.~\cite{eigen2014depth}, with reprojected LIDAR point clouds, and those using the newer and improved ground truth~\cite{uhrig2017sparsity}, which aggregates $5$ consecutive frames and handles dynamic objects. Given the higher reliability of the new ground truth, we used it to evaluate all the baselines on the Eigen test split with all the images that contain ground truth, a total of $652$.
We provide evaluation without and with the Eigen cropping, see Table~\ref{tab:kitti_new_eigen_nocrop} and Table~\ref{tab:kitti_new_eigen_crop}.
For fairness and completeness, as some methods present results with the old ground truth, we also include an evaluation with the LiDAR reprojected depths, on the complete Eigen split with $697$ images. We report additional results directly taken from the corresponding papers, see Table~\ref{tab::kitti_old_eigen_crop}.

In a few of the KITTI test scenes the camera motion is insufficient for proper SfM convergence. Our SfM-TTR cannot refine the depth in those cases, but for a fair comparison, we included these sequences in the global metrics using the results of the network without SfM-TTR.

Note that although we have presented a novel scale alignment, for the sake of fairness we align the self-supervised predictions and the ground truth with the per-image median, as commonly done~\cite{watson2021temporal, godard2019digging}. Also following the common evaluation practices, we set a maximum depth of 80 meters.

\subsection{Comparisons against Baselines}

We demonstrate the benefits of our method by comparing the results of applying a photometric refinement (Ph-TTR) and ours (SfM-TTR) on the baseline networks. Table~\ref{tab:kitti_new_eigen_nocrop} shows how our SfM-TTR consistently and significantly improves the predictions of all networks, obtaining superior performance than the photometric refinement. Besides, Ph-TTR fails to improve over CADepth without TTR. The most likely reason is that it requires individual hyperparameter tuning, which was not required for our SfM-TTR.

\begin{figure}[t!]
    \centering
    \includegraphics[width=0.95\linewidth]{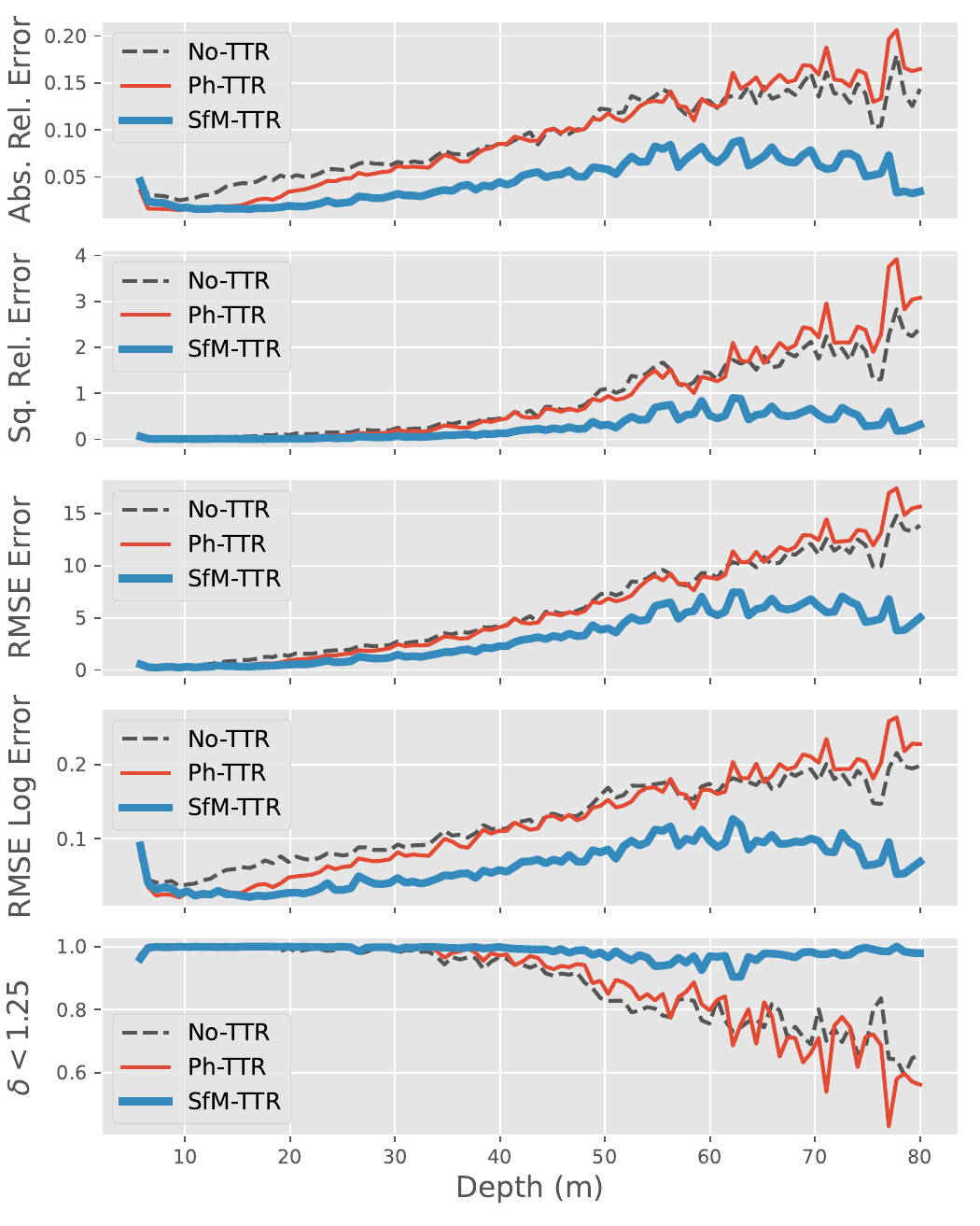}
    \caption{\textbf{Error metrics for different depths with DIFFNet. }Our SfM-TTR (thick blue) gives a substantial improvement over No-TTR (dashed black) and Ph-TTR (thin red) at medium and large depths. Ph-TTR offers some improvement over No-TTR at close depths, where the small baselines of photometric losses are informative, but it does not improve or it is slightly worse at medium and large depths. The metrics $\delta < 1.25^2$ and $\delta < 1.25^3$ are not plotted, as differences are small (see for example Table \ref{tab:kitti_new_eigen_nocrop}).}
    \label{fig:error_vs_depth}
    \vspace{-3mm}
\end{figure}

The advantages of our proposed method are especially noticeable for large depths, where Ph-TTR cannot provide a good supervision signal due to the limited parallax between close frames.
Our refinement, instead, leverages SfM, which triangulates points from the complete sequence. This produces better estimates for distant points and better supervision, resulting in a drastic reduction of the RMSE by up to 30\%. 
This effect is clearly visible in Figure~\ref{fig:error_vs_depth}. Although smaller depths show comparable performance for Ph-TTR and SfM-TTR, the photometric loss does not help in areas with large depths. SfM-TTR, instead, provides a significant gain in performance in those areas.

The best results are obtained when applying our SfM-TTR to DIFFNet, even though the original DIFFNet without TTR performs slightly worse than ManyDepth. We believe that our TTR has a smaller effect on ManyDepth because it already leverages scene information by using multiple frames at inference time. SfM-TTR can also improve results on AdaBins, for which Ph-TTR cannot be implemented, as AdaBins does not provide a pose estimation module. This further demonstrates the effectiveness of directly optimizing for the 3D points from COLMAP.

\begin{figure}
    \centering
    \includegraphics[width=0.95\linewidth]{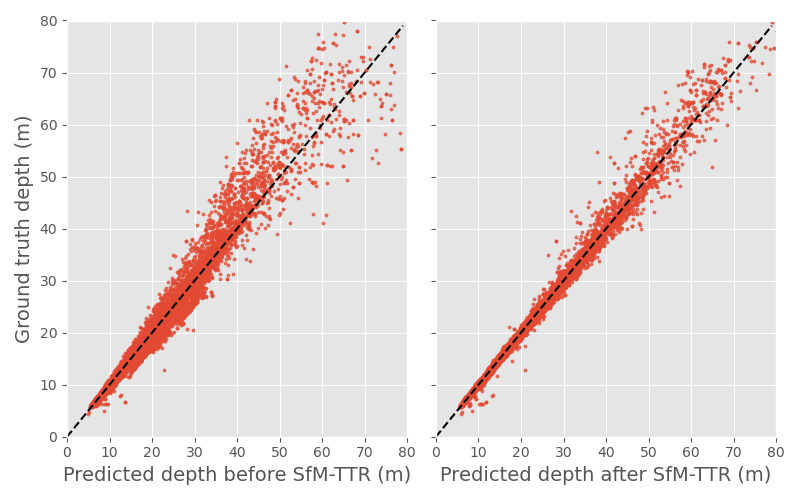}
    \caption{\textbf{Depth predictions and ground truth before and after SfM-TTR with DIFFNet.} The red dots stand for predicted pixel depths on a KITTI sequence with DIFFNet, the black dashed line stands for zero error. Note how after SfM-TTR the red dots gather closer to the dashed black line, illustrating that the predicted depths are closer to the ground truth ones.}
    \label{fig:prev_post_scatter}
    \vspace{-2mm}
\end{figure}

\begin{figure*}[!t]
\centering

\newcommand{\customcolumnwidth}{0.3\textwidth}

\resizebox{1.0\textwidth}{!}{\begin{tabular}{@{\hspace{1.5mm}}c@{\hspace{1.5mm}}c@{\hskip 2mm}c@{\hspace{1.5mm}}c@{\hspace{1.5mm}}c@{}}

& No TTR & Ph-TTR & \textbf{SfM-TTR} \\

{\rotatebox{90}{\scriptsize AdaBins}} &
\includegraphics[width=\customcolumnwidth]{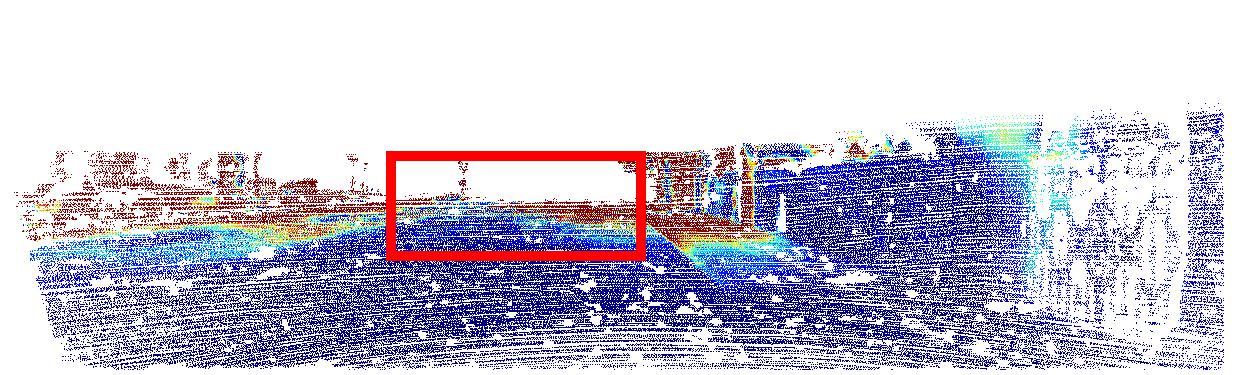} &
\includegraphics[width=\customcolumnwidth]{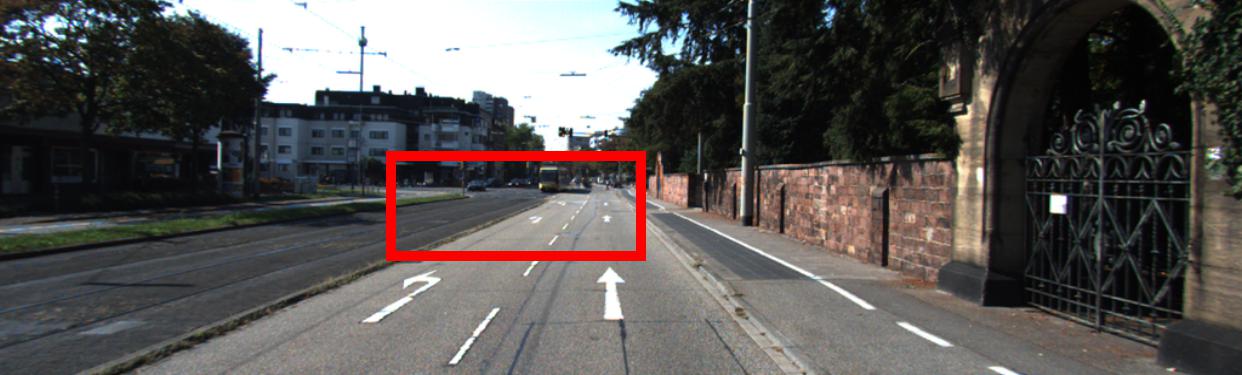} &
\includegraphics[width=\customcolumnwidth]{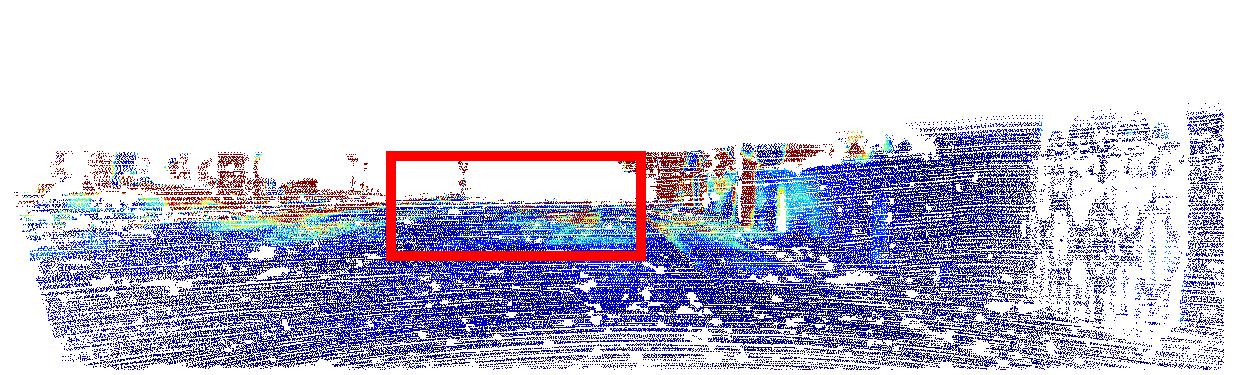} \\[-3mm]

{\rotatebox{90}{\scriptsize ManyDepth}} &
\includegraphics[width=\customcolumnwidth]{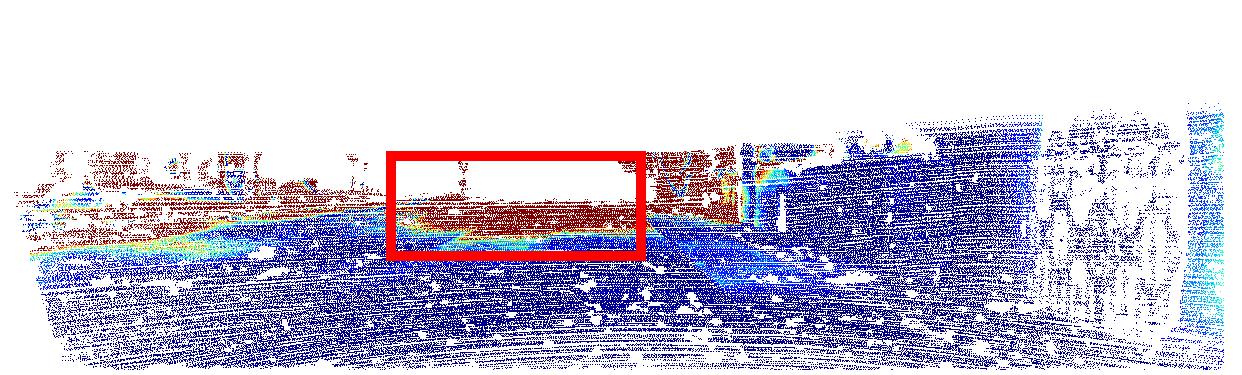} &
\includegraphics[width=\customcolumnwidth]{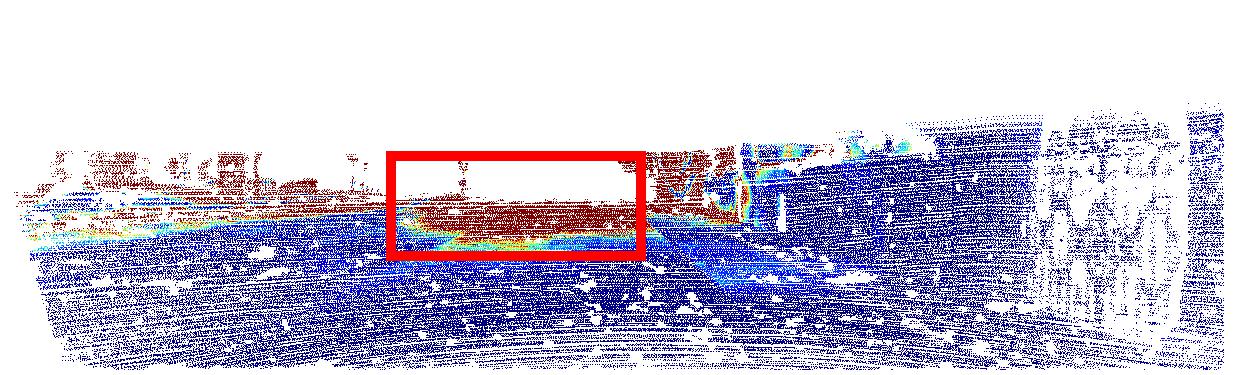} &
\includegraphics[width=\customcolumnwidth]{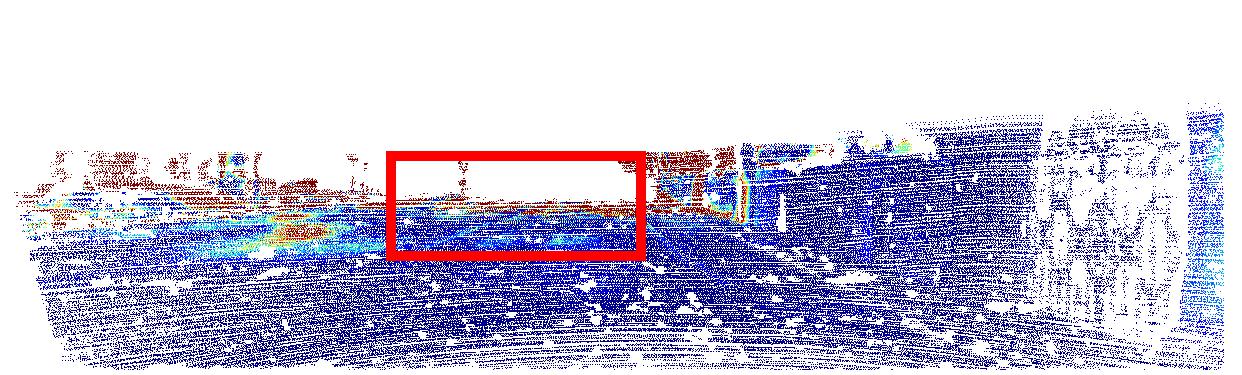} \\[-3mm]

{\rotatebox{90}{\scriptsize CADepth}} &
\includegraphics[width=\customcolumnwidth]{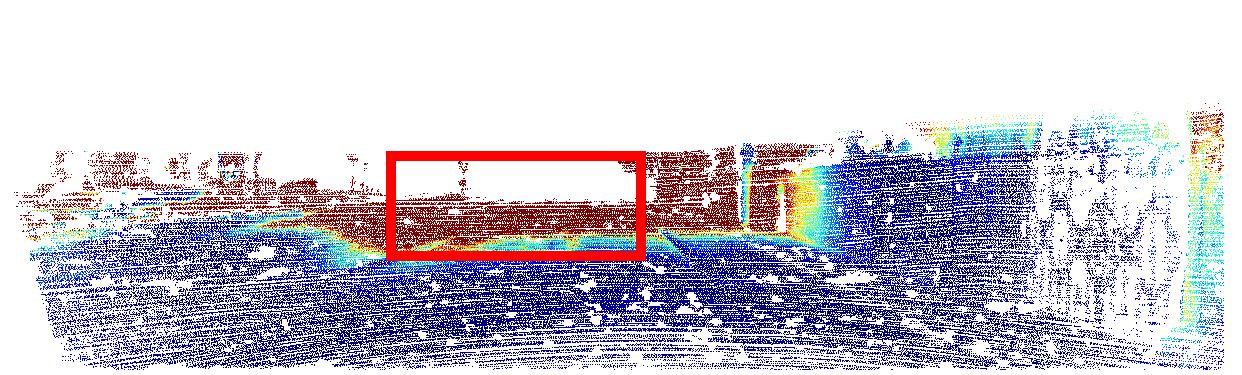} &
\includegraphics[width=\customcolumnwidth]{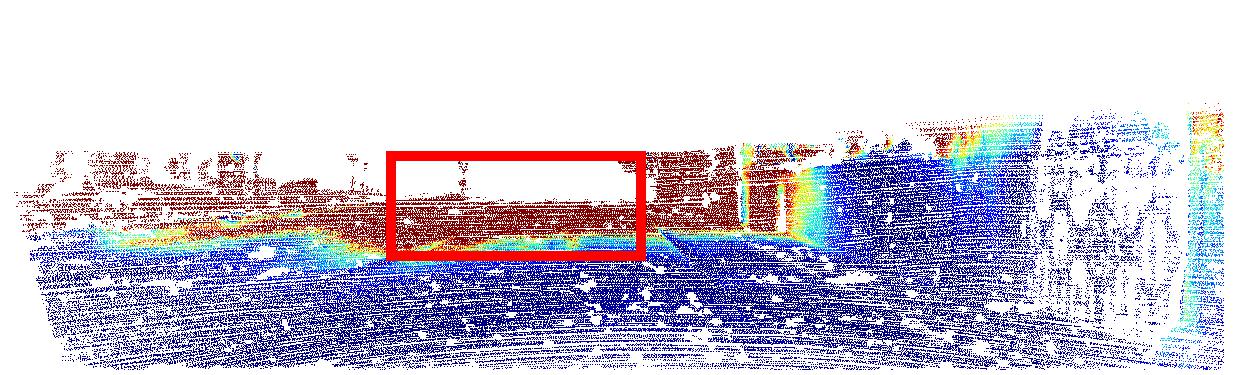} &
\includegraphics[width=\customcolumnwidth]{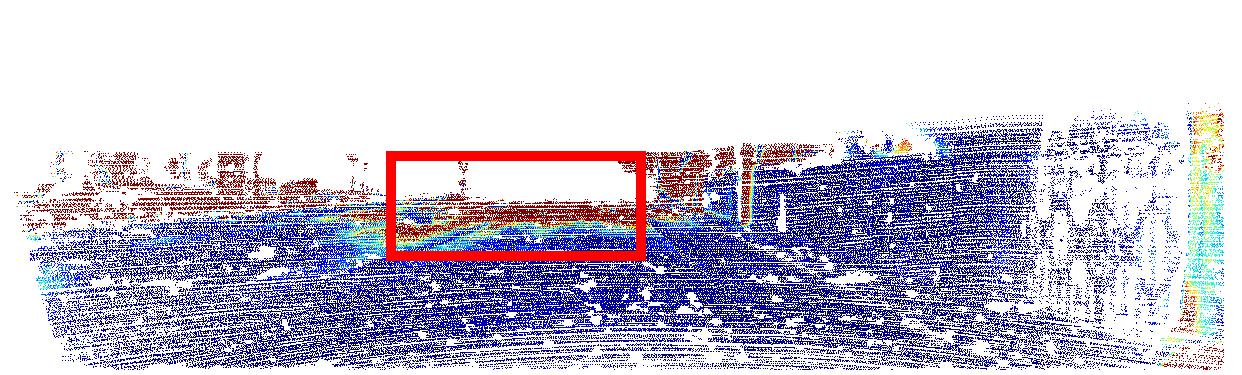} \\[-3mm]

{\rotatebox{90}{\scriptsize DIFFNet}} &
\includegraphics[width=\customcolumnwidth]{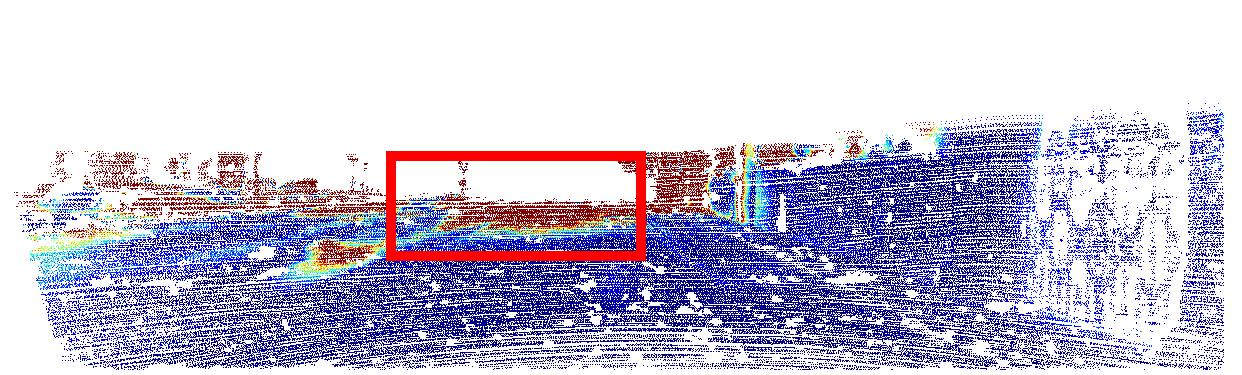} &
\includegraphics[width=\customcolumnwidth]{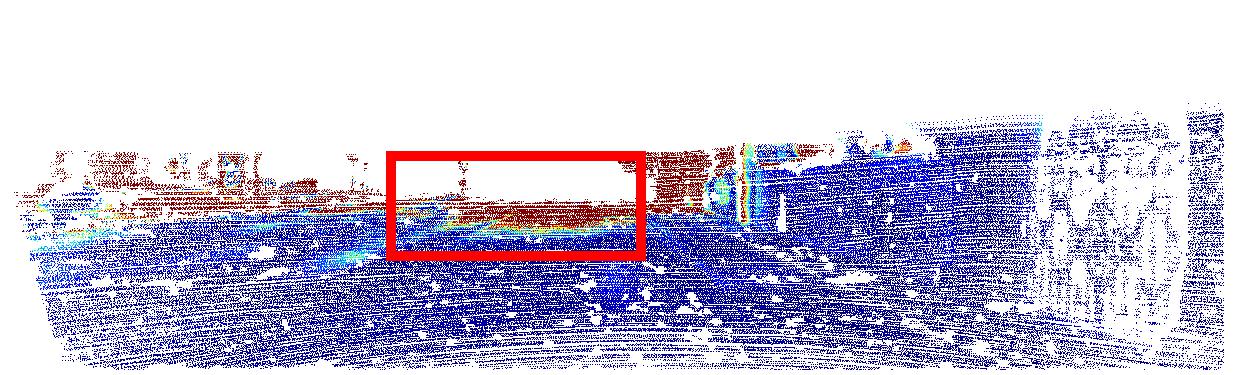} &
\includegraphics[width=\customcolumnwidth]{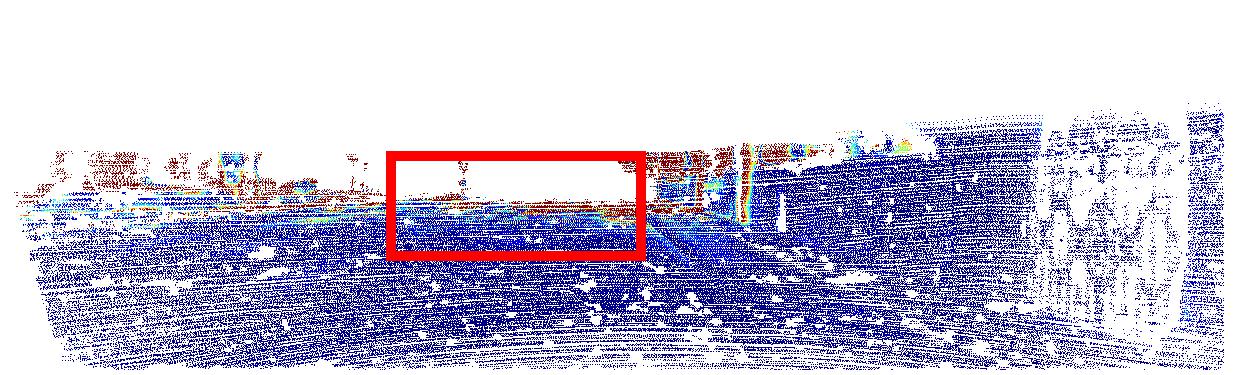} \\

& & & \\

\hline

& & & \\

{\rotatebox{90}{\scriptsize AdaBins}} &
\includegraphics[width=\customcolumnwidth]{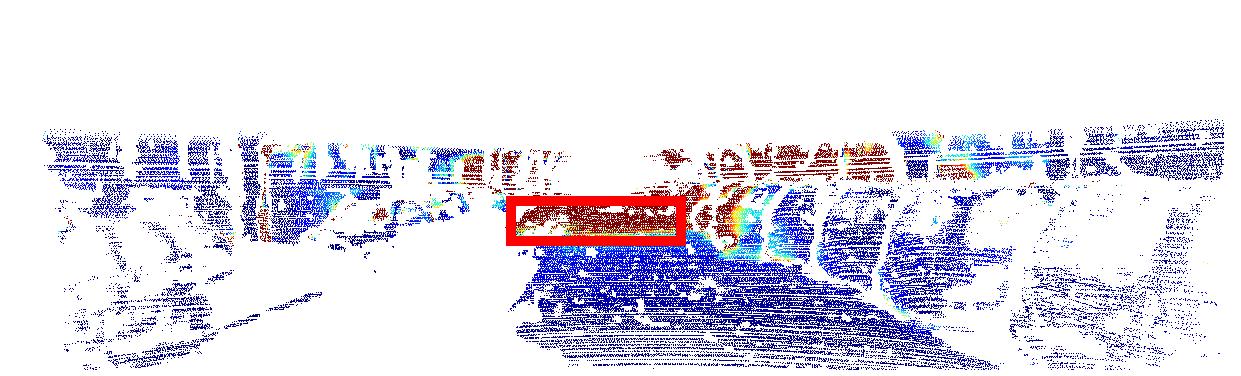} &
\includegraphics[width=\customcolumnwidth]{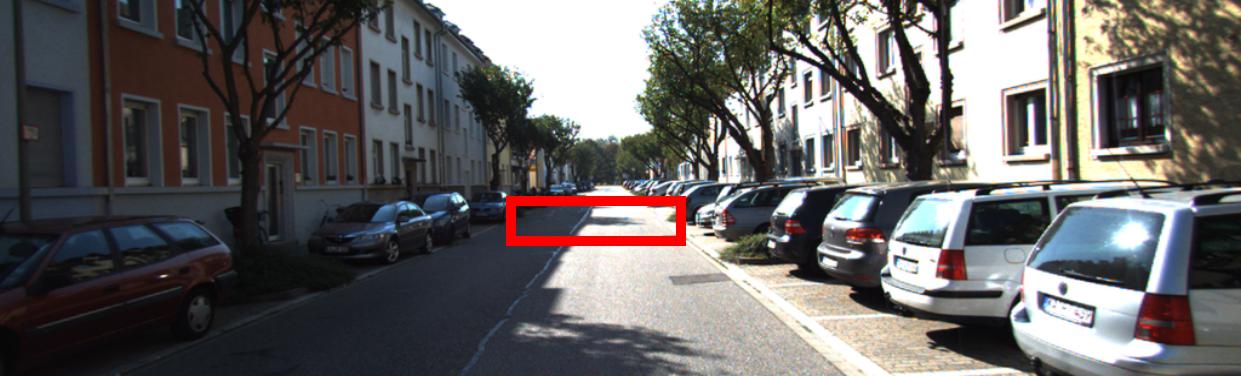} &
\includegraphics[width=\customcolumnwidth]{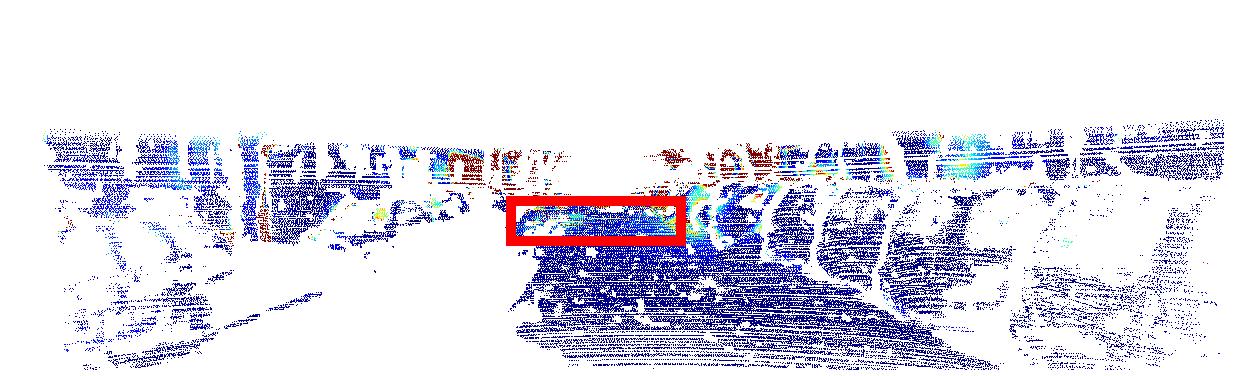} \\[-3mm]

{\rotatebox{90}{\scriptsize ManyDepth}} &
\includegraphics[width=\customcolumnwidth]{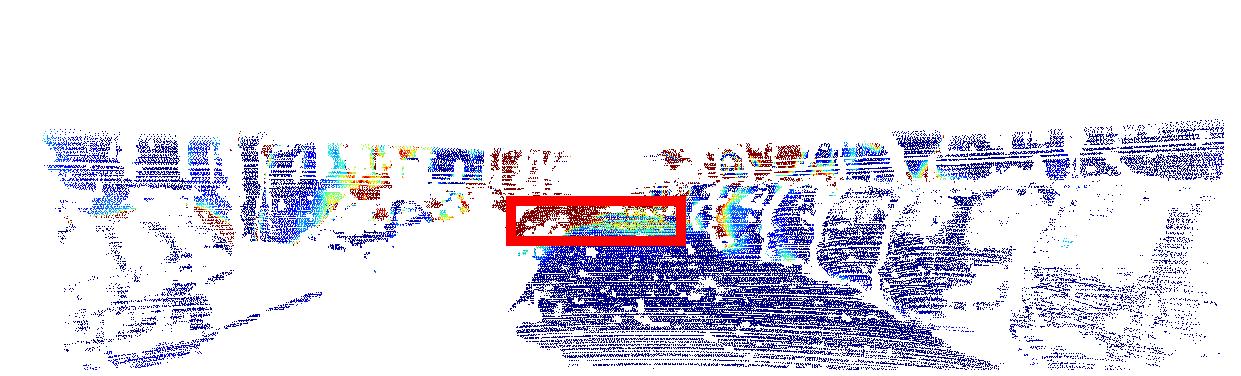} &
\includegraphics[width=\customcolumnwidth]{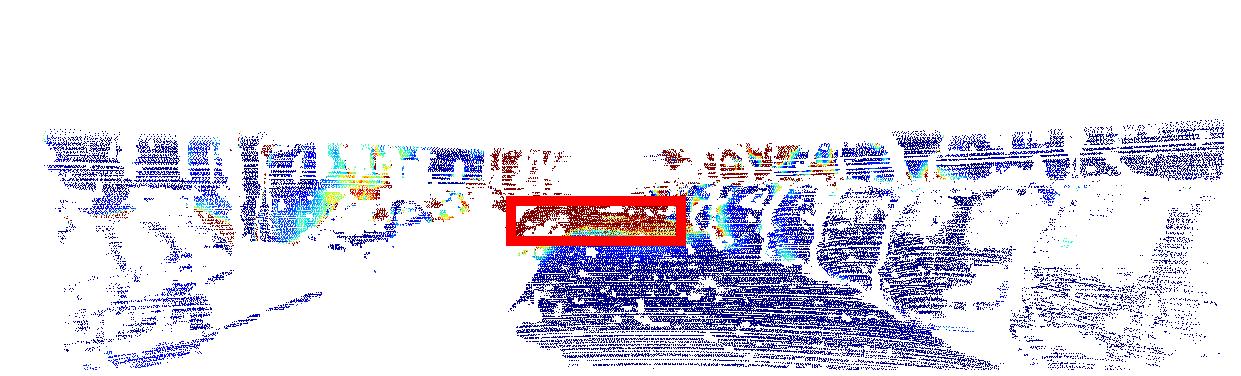} &
\includegraphics[width=\customcolumnwidth]{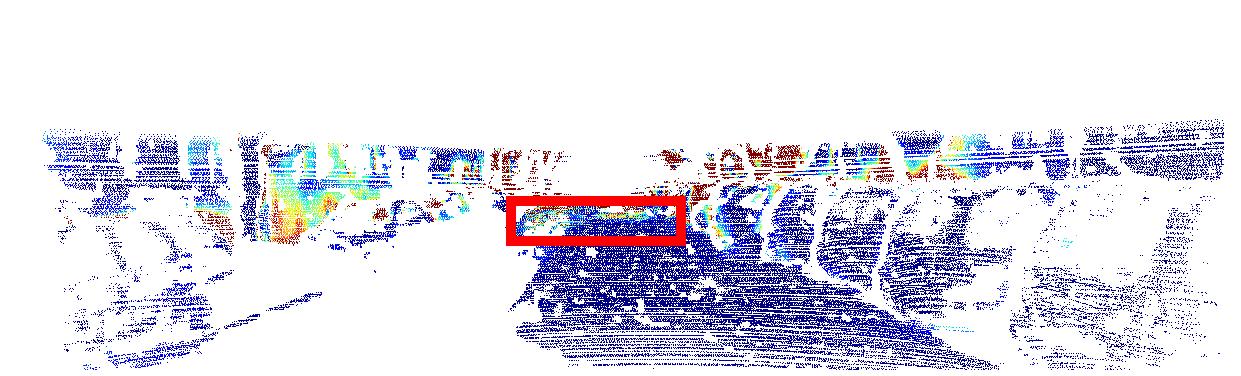} \\[-3mm]

{\rotatebox{90}{\scriptsize CADepth}} &
\includegraphics[width=\customcolumnwidth]{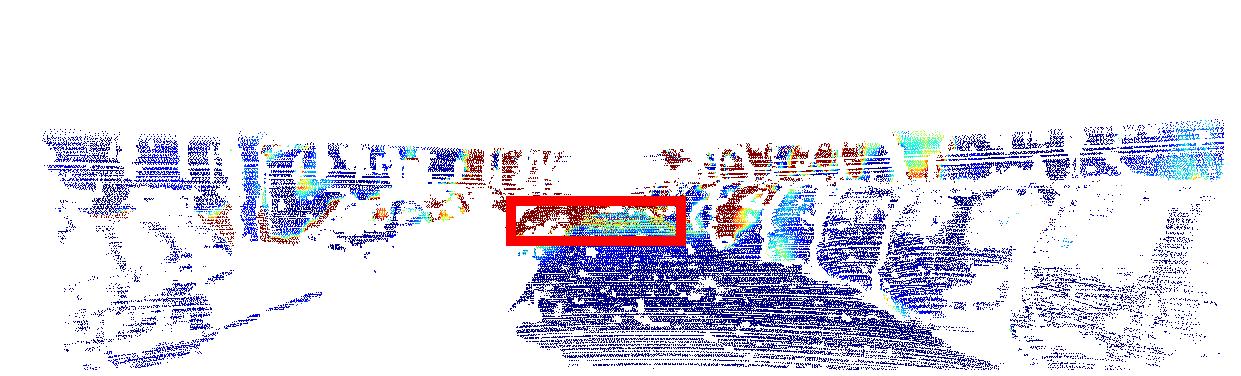} &
\includegraphics[width=\customcolumnwidth]{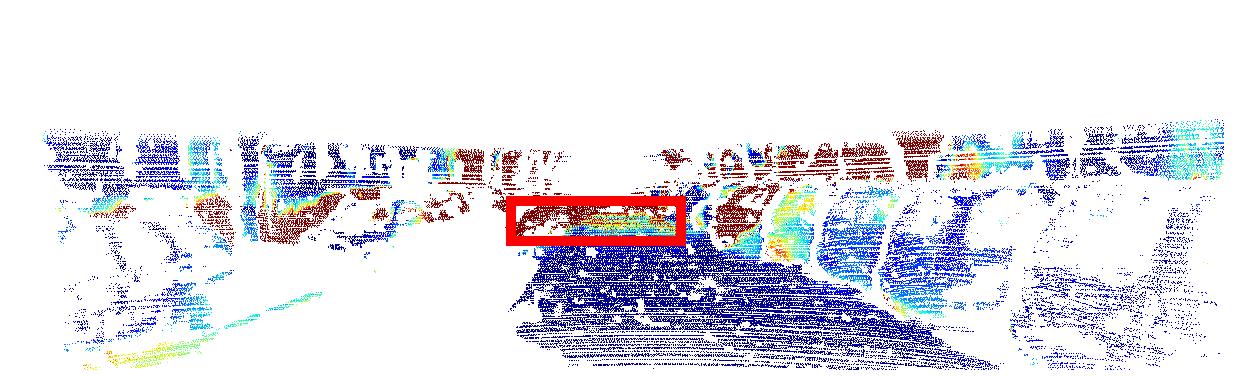} &
\includegraphics[width=\customcolumnwidth]{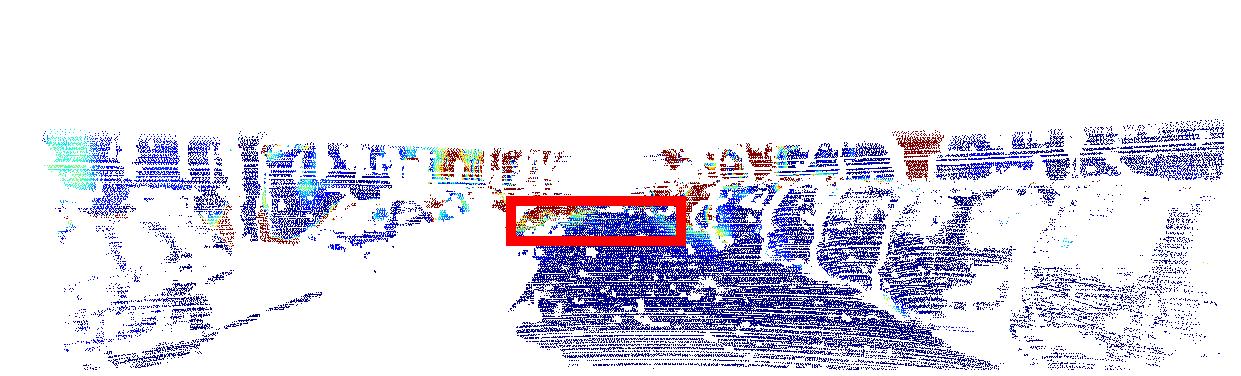} \\[-3mm]

{\rotatebox{90}{\scriptsize DIFFNet}} &
\includegraphics[width=\customcolumnwidth]{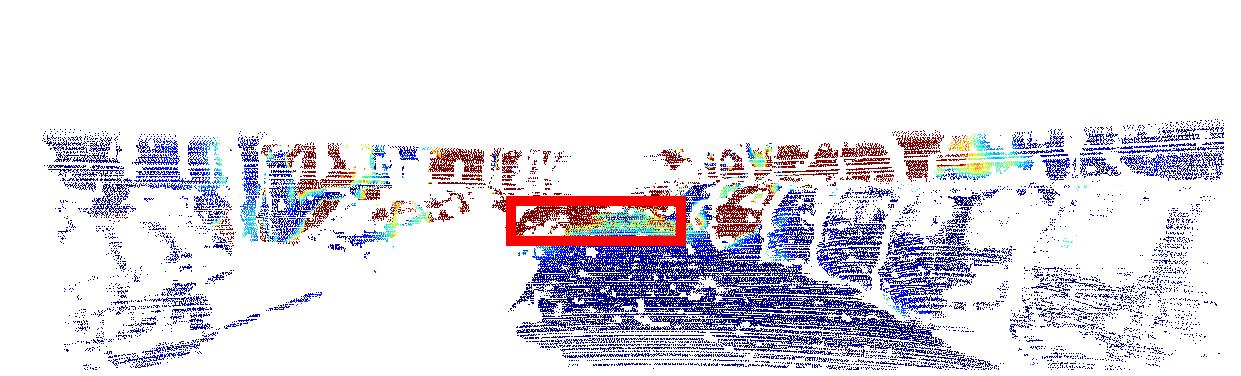} &
\includegraphics[width=\customcolumnwidth]{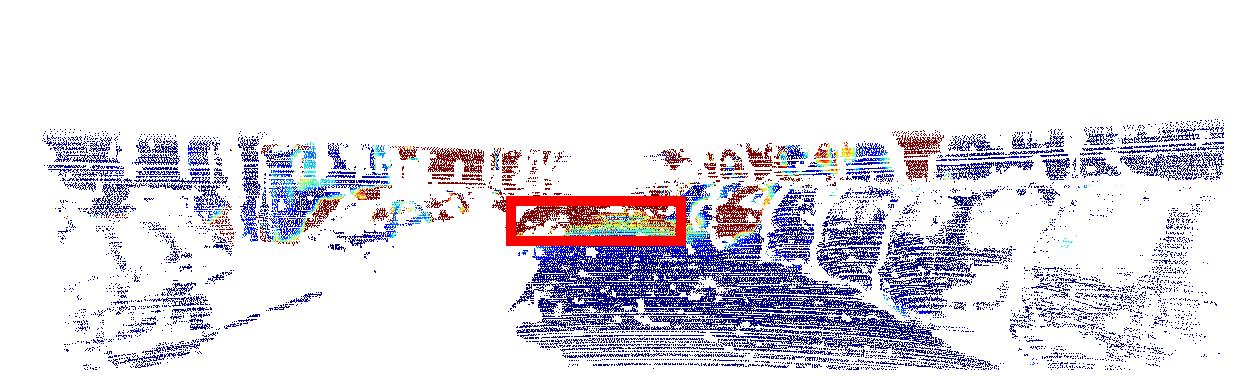} &
\includegraphics[width=\customcolumnwidth]{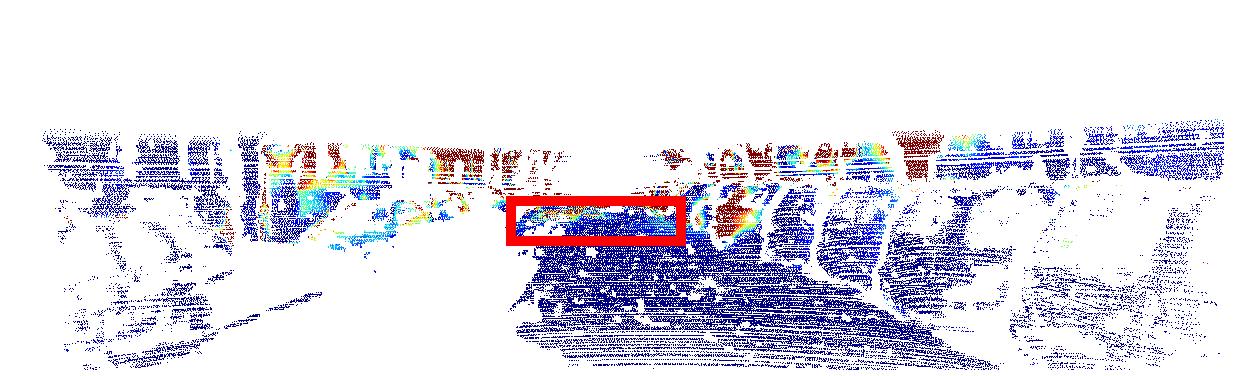} \\

\end{tabular}}

\caption{\textbf{RMSE maps for different baselines architectures (rows) and TTR (columns).} The input image is the center top image, as AdaBins cannot be refined with photometric loss. The benefit of our SfM-TTR is particularly noticeable for large depths (framed by red rectangles). Ph-TTR methods struggle in these areas as they use weak low-parallax constraints, while SfM leverages wider baselines and produces more accurate depth supervision. Figure best viewed in color.}

\label{fig:error_maps} 
\end{figure*}

\begin{figure*}[!t]
\centering

\newcommand{\customcolumnwidth}{0.19\textwidth}

\begin{tabular}{@{\hspace{1.5mm}}c@{\hspace{1.5mm}}c@{\hspace{1.5mm}}c@{\hspace{1.5mm}}c@{\hspace{1.5mm}}c@{}}

Input & AdaBins + SfM-TTR & ManyDepth + SfM-TTR & CADepth + SfM-TTR & DIFFNet + SfM-TTR \\

\includegraphics[width=\customcolumnwidth]{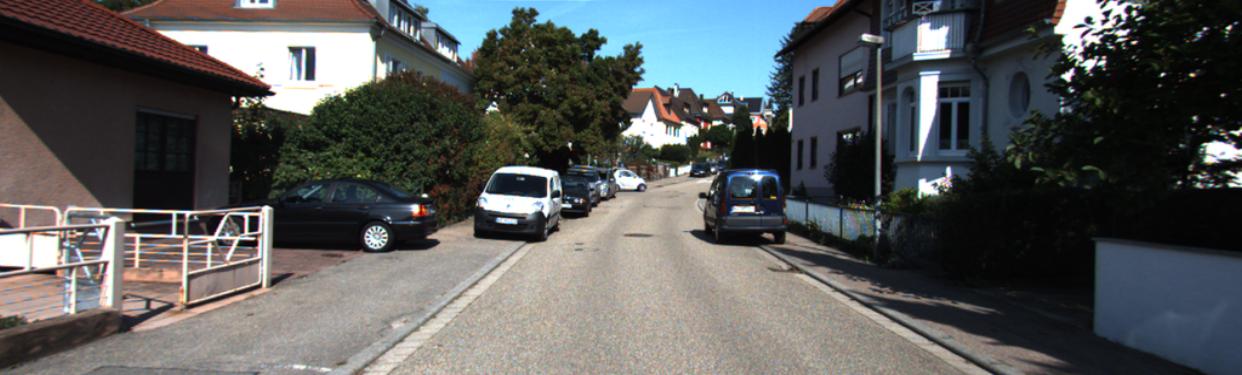} &
\includegraphics[width=\customcolumnwidth]{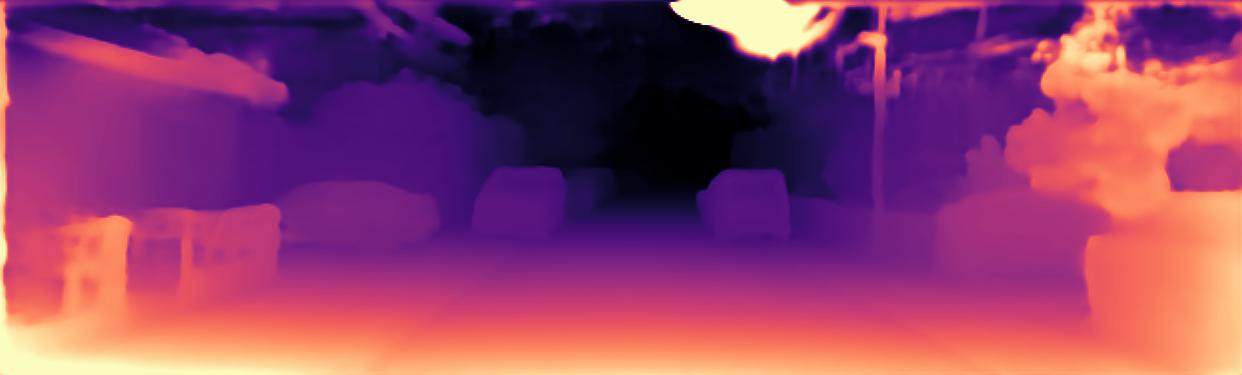} &
\includegraphics[width=\customcolumnwidth]{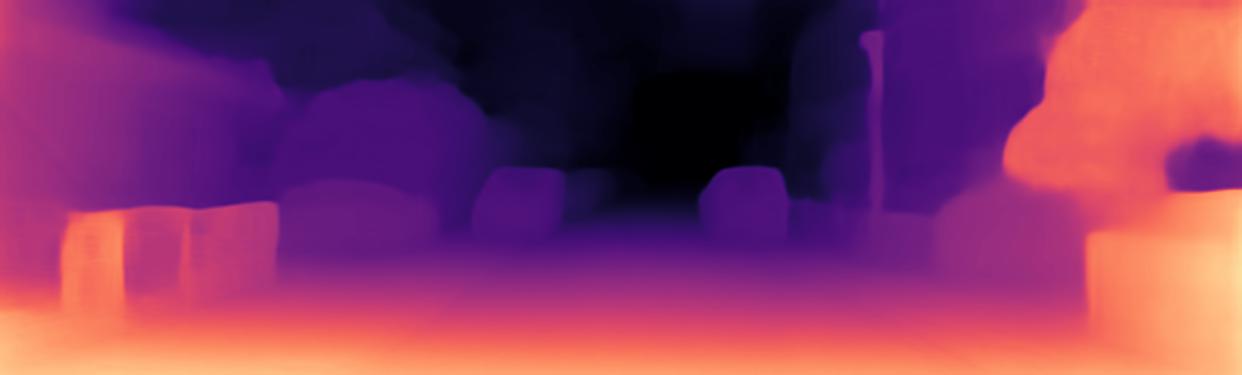} &
\includegraphics[width=\customcolumnwidth]{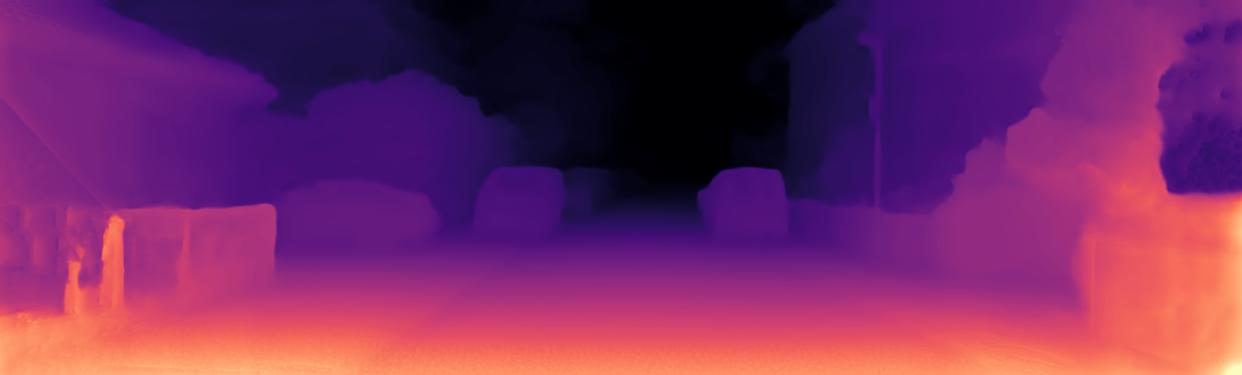} &
\includegraphics[width=\customcolumnwidth]{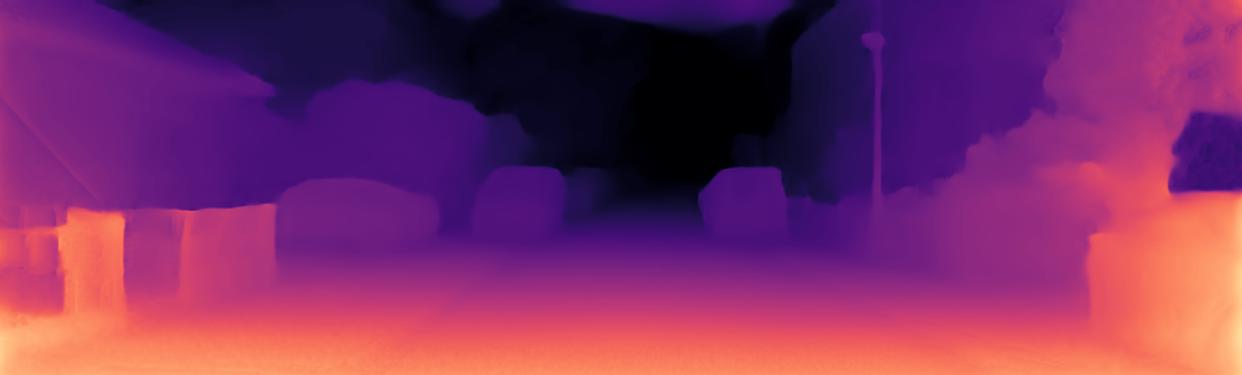} \\

\includegraphics[width=\customcolumnwidth]{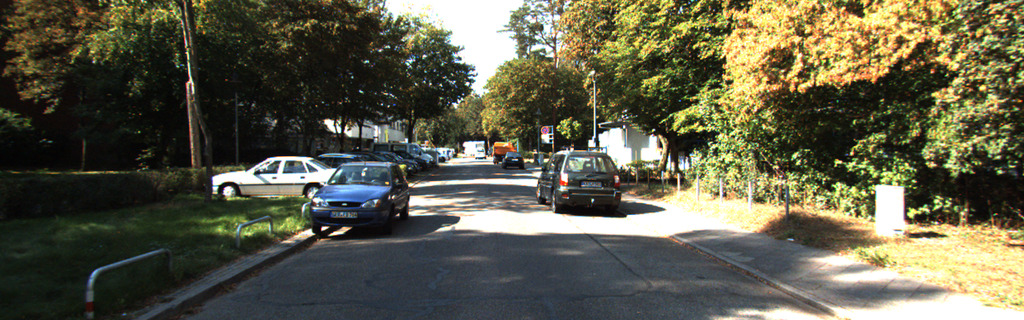} &
\includegraphics[width=\customcolumnwidth]{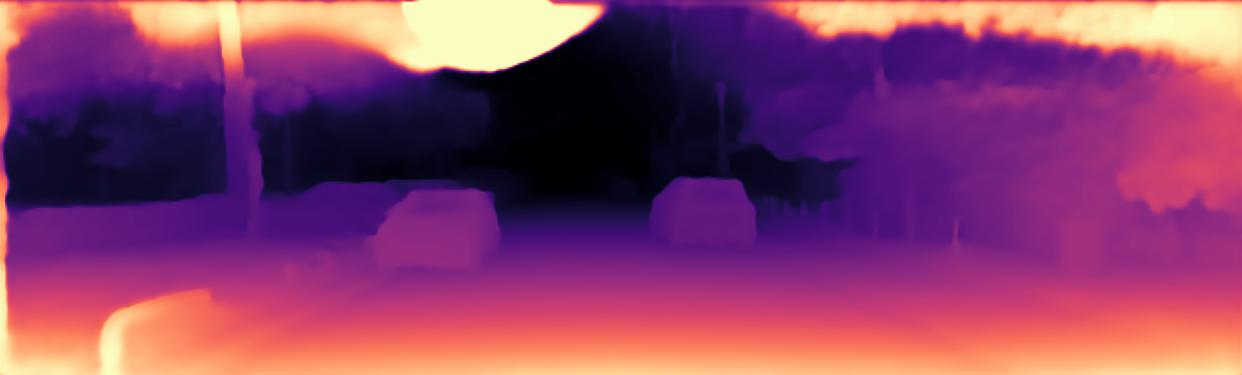} &
\includegraphics[width=\customcolumnwidth]{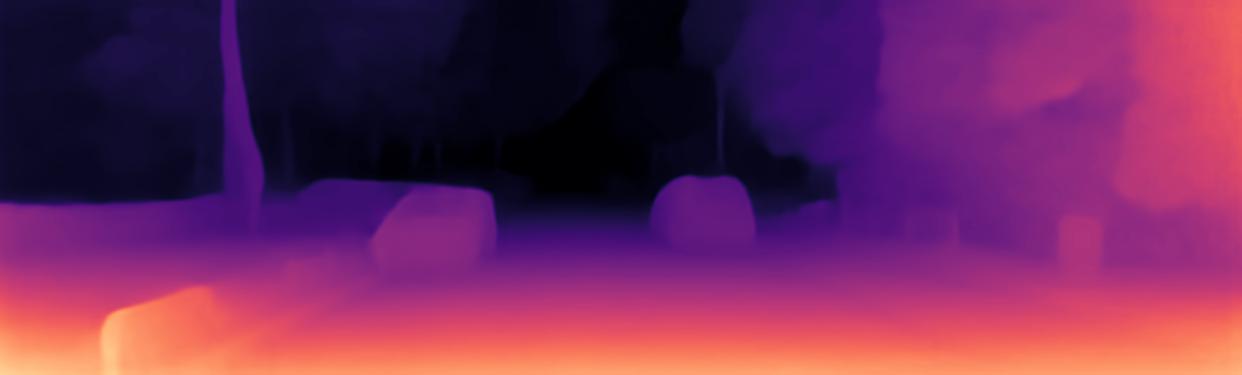} &
\includegraphics[width=\customcolumnwidth]{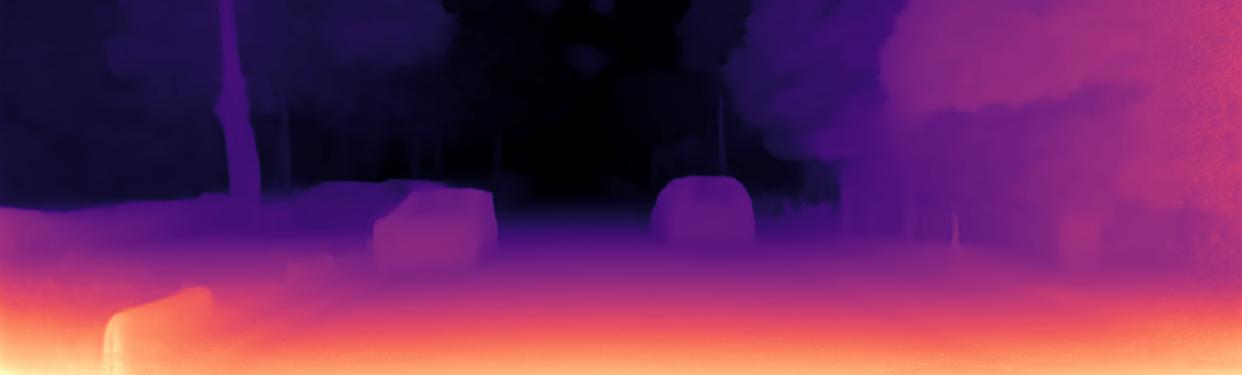} &
\includegraphics[width=\customcolumnwidth]{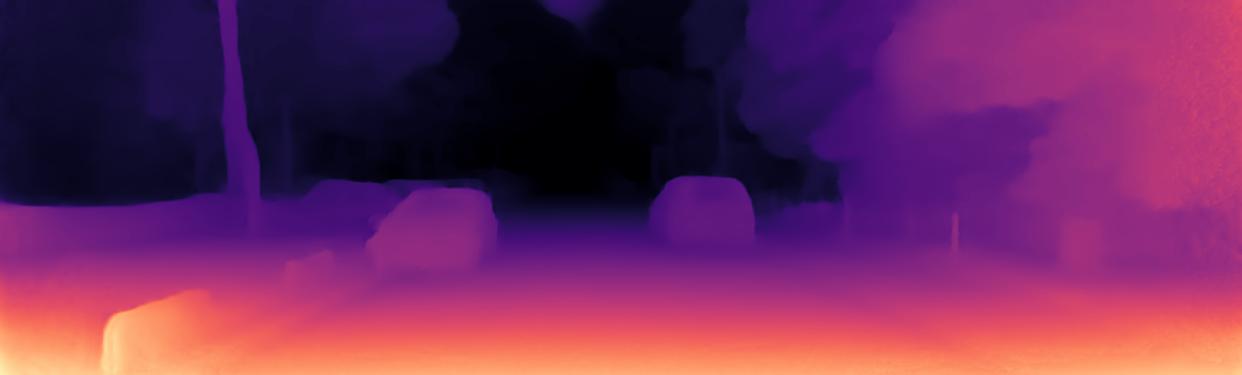} \\

\includegraphics[width=\customcolumnwidth]{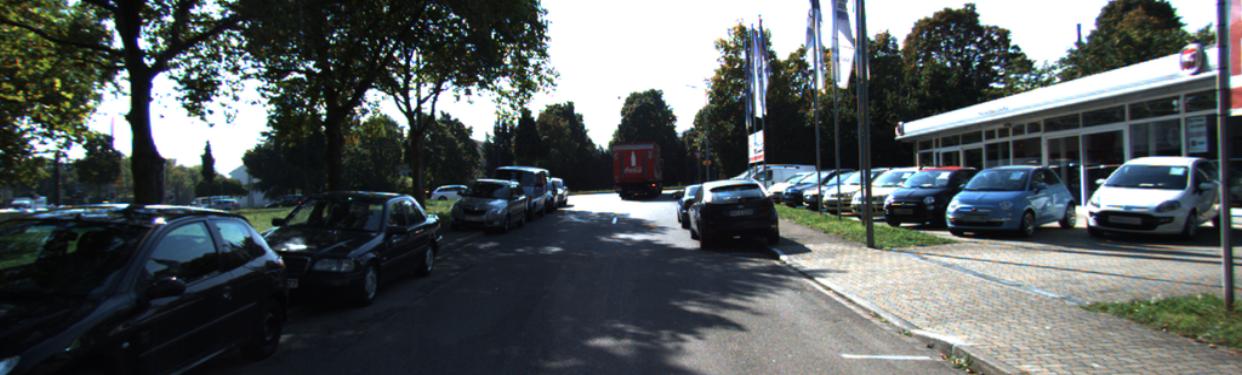} &
\includegraphics[width=\customcolumnwidth]{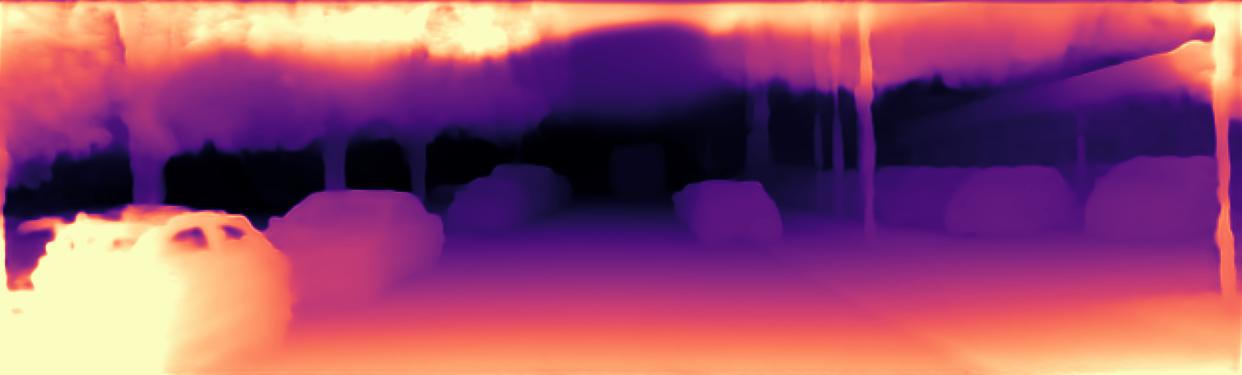} &
\includegraphics[width=\customcolumnwidth]{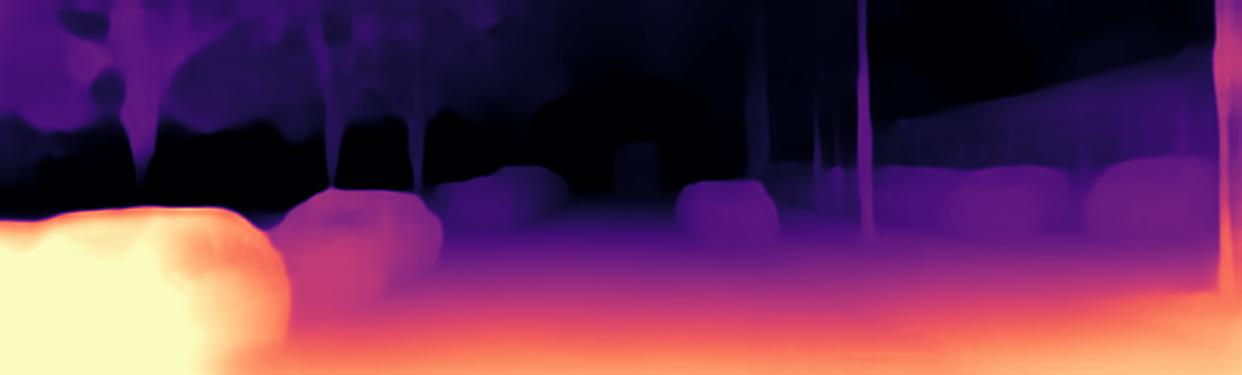} &
\includegraphics[width=\customcolumnwidth]{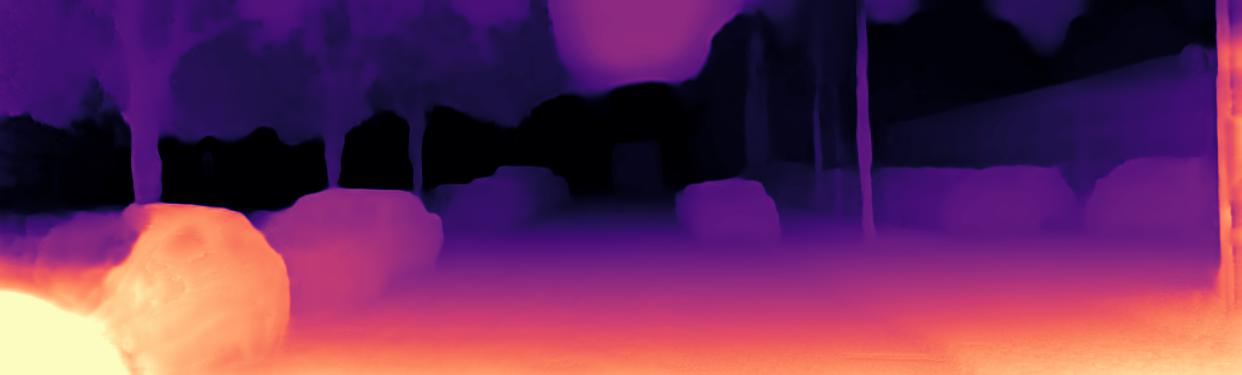} &
\includegraphics[width=\customcolumnwidth]{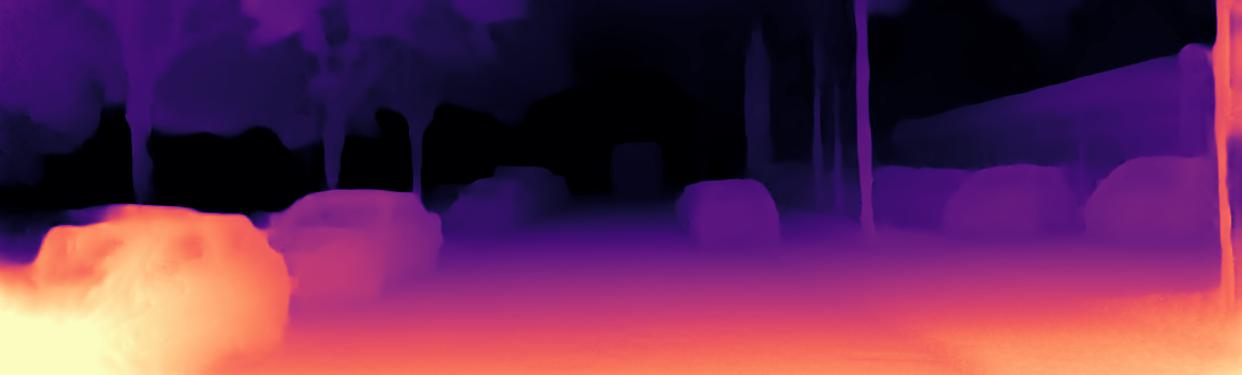} \\

\includegraphics[width=\customcolumnwidth]{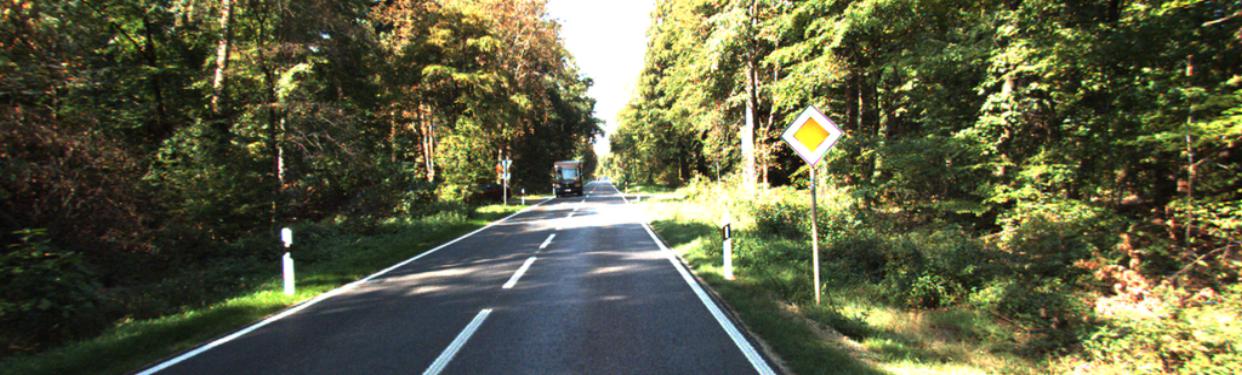} &
\includegraphics[width=\customcolumnwidth]{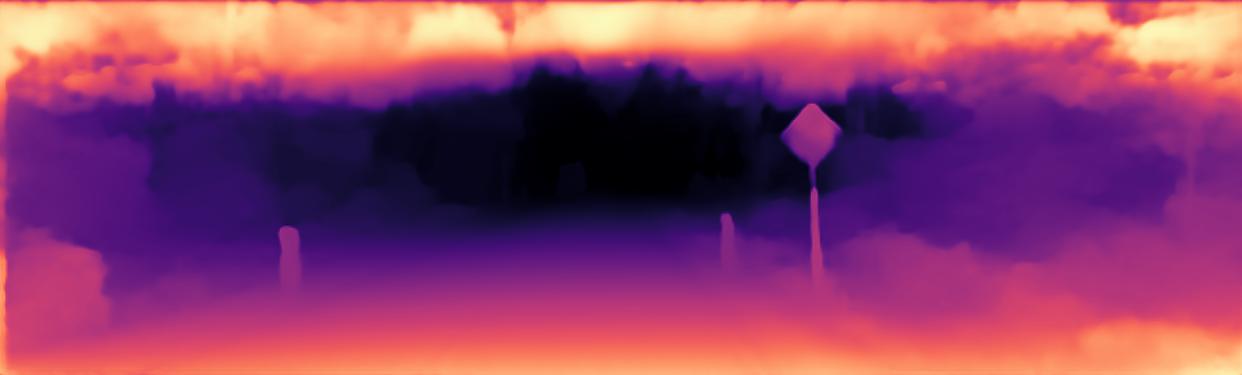} &
\includegraphics[width=\customcolumnwidth]{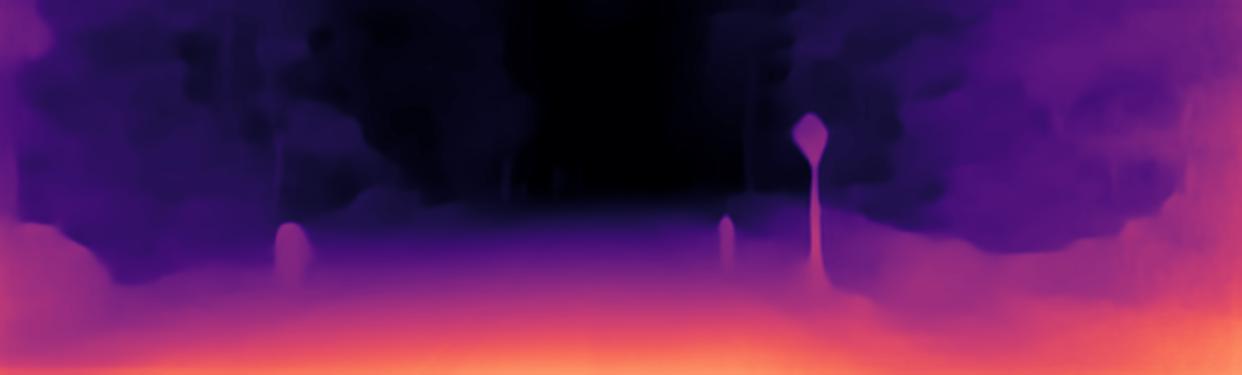} &
\includegraphics[width=\customcolumnwidth]{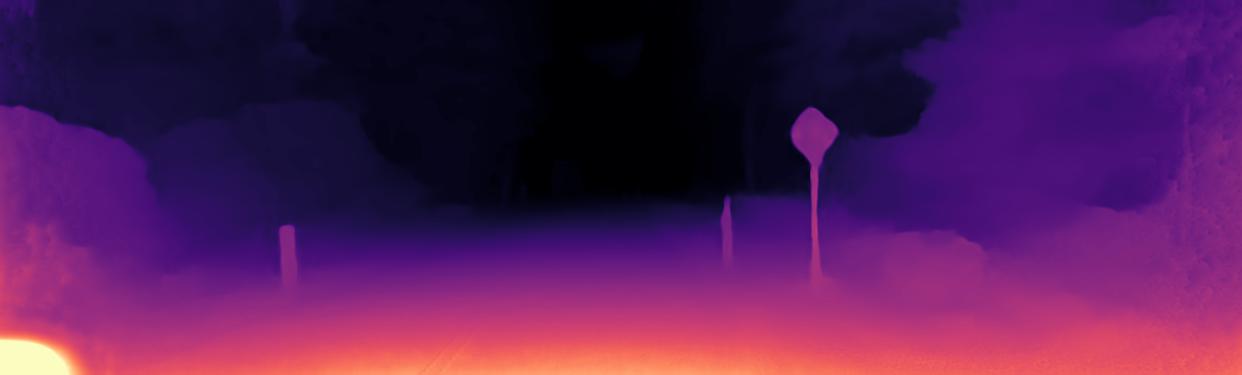} &
\includegraphics[width=\customcolumnwidth]{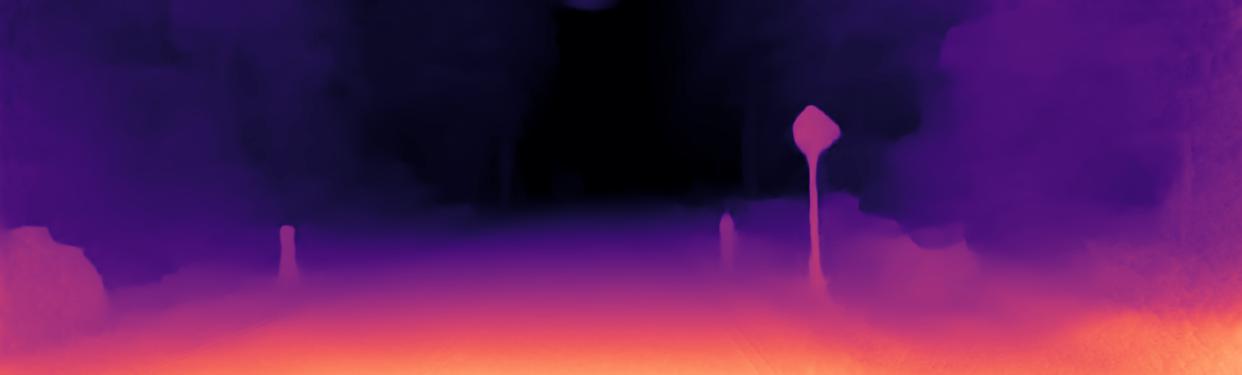} \\

\end{tabular}

\caption{{Qualitative depth maps for different architectures after SfM-TTR on KITTI.}}
\label{fig:qualitative_kitti} 
\end{figure*}

Qualitatively, Figure~\ref{fig:qualitative_kitti} shows how predictions after SfM-TTR keep looking sharp with well-defined boundaries despite the sparsity of the pseudo-ground truth. We argue that optimizing the encoder enables a better understanding of the scene while freezing the decoder maintains the previously learned sharpness of the predictions. The error maps from Figure~\ref{fig:error_maps} reveal the differences between refinements, showing how our method can effectively reduce errors in regions where Ph-TTR cannot. The positive effect of SfM-TTR in distant points is visible in Figure~\ref{fig:prev_post_scatter}, where large depths move closer to the ground truth after our refinement.

Regarding runtime efficiency, our method requires roughly 2 seconds per frame during the optimization, similar to Ph-TTR, and faster than other multi-view TTR that also use large baselines~\cite{tiwari2020pseudo, luo2020consistent}.

\subsection{Ablation Studies}

To validate the relative importance of the individual components of our SfM-TTR, we perform ablation studies where we dispose some of our key components.

\begin{table}[t]
  \centering
  \footnotesize
  \resizebox{1.0\linewidth}{!}{
    \begin{tabular}{|l|c|c|c|c|}\hline
        Method & Abs Rel~$\downarrow$ & Sq Rel~$\downarrow$ & RMSE~$\downarrow$ & RMSE log~$\downarrow$ \\
    \hline\hline

AdaBins~\cite{bhat2021adabins}                               &         0.072  &         0.325  &         3.134  &         0.112  \\
AdaBins~\cite{bhat2021adabins} + SfM-TTR (full model)        &         0.062  & \textbf{0.204} &         2.297  &         0.092  \\
\textbf{AdaBins~\cite{bhat2021adabins} + SfM-TTR (encoder)}  & \textbf{0.060} & \textbf{0.204} & \textbf{2.260} & \textbf{0.091} \\
    \hline
ManyDepth~\cite{watson2021temporal}                          &         0.064  &         0.345  &         3.116  &         0.103  \\
ManyDepth~\cite{watson2021temporal} + SfM-TTR (full model)   &         0.059  & \textbf{0.293} &         2.655  &         0.096  \\
\textbf{ManyDepth~\cite{watson2021temporal} + SfM-TTR (encoder)} & \textbf{0.057} &         0.294  & \textbf{2.648} & \textbf{0.094} \\
    \hline
CADepth~\cite{yan2021channel}                                &         0.078  &         0.403  &         3.432  &         0.119  \\
CADepth~\cite{yan2021channel} + SfM-TTR (full model)         &         0.069  & \textbf{0.321} &         2.824  & \textbf{0.104} \\
\textbf{CADepth~\cite{yan2021channel} + SfM-TTR (encoder)}   & \textbf{0.068} &         0.328  & \textbf{2.821} &         0.106  \\
    \hline
DIFFNet~\cite{zhou2021diffnet}                               &         0.071  &         0.361  &         3.230  &         0.110  \\
DIFFNet~\cite{zhou2021diffnet} + SfM-TTR (full model)        &         0.057  & \textbf{0.273} &         2.621  & \textbf{0.092} \\
\textbf{DIFFNet~\cite{zhou2021diffnet} + SfM-TTR (encoder)}  & \textbf{0.056} & \textbf{0.273} & \textbf{2.600} &         0.093  \\
    \hline

         \hline
    \end{tabular}
  }
  \caption{\textbf{Encoder vs. full network TTR.} Note how the best results are achieved with encoder-only TTR.}
  \label{tab:ablation_1}
  \vspace{-2mm}
\end{table}

Table~\ref{tab:ablation_1} shows a comparison between refining the complete network and only updating the encoder. Similar to~\cite{mccraith2020monocular}, we obtain better results when only updating the encoder, further showing how light refinement schemes should only focus on improving the underlying representation of the network.

\begin{table}[t]
  \centering
  \footnotesize
  \resizebox{\linewidth}{!}{
    \begin{tabular}{|l|c|c|c|c|}\hline
        Method & Abs Rel~$\downarrow$ & Sq Rel~$\downarrow$ & RMSE~$\downarrow$ & RMSE log~$\downarrow$ \\
    \hline\hline
AdaBins~\cite{bhat2021adabins}                               &         0.072  &         0.325  &         3.134  &         0.112  \\
AdaBins~\cite{bhat2021adabins} + SfM-TTR (median)            &         0.074  &         0.263  &         2.509  &         0.103  \\
AdaBins~\cite{bhat2021adabins} + SfM-TTR ($\bfdepthnetscheck, \bfdepthsfmcheck$)            &         0.065  &         0.278  &         2.787  &         0.103  \\
AdaBins~\cite{bhat2021adabins} + SfM-TTR (Least Squares)     &         0.064  &         0.222  &         2.346  &         0.097  \\
AdaBins~\cite{bhat2021adabins} + SfM-TTR ($w_{l,j}^{\boldsymbol{\theta}}=1$)             &         0.062  &         0.206  &         2.310  & \textbf{0.091}  \\
\textbf{AdaBins~\cite{bhat2021adabins} + SfM-TTR}            & \textbf{0.060} & \textbf{0.204} & \textbf{2.260} & \textbf{0.091} \\
    \hline
    \end{tabular}
    }
  \caption{\textbf{Alignment ablation study.} Note the substantial improvement of our scaling approach (detailed in Section \ref{sec:scaling}) over other alignments.} 
  \label{tab:ablation_2}
  \vspace{-3mm}
\end{table}

As shown in Table~\ref{tab:ablation_2}, using the mean of per-image medians~\cite{luo2020consistent, klodt2018supervising} alignment in our SfM-TTR, as well as other ablated versions of our method, worsens significantly the performance on \mbox{AdaBins}.
The alignment is specially important for supervised models, as their scale is not corrected during the evaluation.
With our alignment, we are accounting for outliers with RANSAC and for the heteroscedastic nature of the depth noise with weighted least squares, resulting in substantially more robust and accurate results.

\begin{table*}[t]
  \centering
  \footnotesize
  \resizebox{1.0\textwidth}{!}{
    \begin{tabular}{|c|l|c|c|c|c|c|c|c|}\hline
        TTR & Method & Abs Rel~$\downarrow$ & Sq Rel~$\downarrow$ & RMSE~$\downarrow$  & RMSE log~$\downarrow$ & $\delta < 1.25~\uparrow$ & $\delta < 1.25^{2}~\uparrow$ & $\delta < 1.25^{3}~\uparrow$ \\
    \hline\hline

       \xmark  & AdaBins~\cite{bhat2021adabins}\expresult\textdagger                             &         0.072  &         0.325  &         3.134  &         0.112  &         0.941  &         0.990  & \textbf{0.998} \\
       \cmark  & \textbf{AdaBins~\cite{bhat2021adabins} + SfM-TTR}            & \textbf{0.060} & \textbf{0.204} & \textbf{2.260} & \textbf{0.091} & \textbf{0.970} & \textbf{0.993} & \textbf{0.998} \\
    \hline
    \hline
       \xmark  & ManyDepth~\cite{watson2021temporal}\paperresult                          &         0.064  &         0.345  &         3.116  &         0.103  &         0.949  &         0.989  & \textbf{0.997} \\
       \cmark  & ManyDepth~\cite{watson2021temporal} + Ph-TTR\paperresult                 & \underline{\textbf{0.056}} &         0.322  &         3.034  &         0.096  &         0.961  & \underline{\textbf{0.992}} & \textbf{0.997} \\
       \cmark  & \textbf{ManyDepth~\cite{watson2021temporal} + SfM-TTR}       &         0.057  & \textbf{0.294} & \textbf{2.648} & \textbf{0.094} & \textbf{0.963} &         0.990  & \textbf{0.997} \\
    \hline
       \xmark  & CADepth~\cite{yan2021channel}\expresult                                &         0.078  &         0.403  &         3.432  &         0.119  &         0.933  &         0.988  & \textbf{0.997} \\
       \cmark  & CADepth~\cite{yan2021channel} + Ph-TTR\expresult                       &         0.088  &         0.475  &         3.723  &         0.132  &         0.914  &         0.984  &         0.996  \\
       \cmark  & \textbf{CADepth~\cite{yan2021channel} + SfM-TTR}             & \textbf{0.068} & \textbf{0.328} & \textbf{2.821} & \textbf{0.106} & \textbf{0.955} & \textbf{0.990} &         0.996  \\
    \hline
       \xmark  & DIFFNet~\cite{zhou2021diffnet}\expresult                               &         0.071  &         0.361  &         3.230  &         0.110  &         0.946  &         0.990  &         0.997  \\
       \cmark  & DIFFNet~\cite{zhou2021diffnet} + Ph-TTR\expresult                      &         0.057  &         0.285  &         2.900  &         0.095  &         0.961  & \underline{\textbf{0.992}} & \underline{\textbf{0.998}} \\
       \cmark  & \textbf{DIFFNet~\cite{zhou2021diffnet} + SfM-TTR}            & \underline{\textbf{0.056}} & \underline{\textbf{0.273}} & \underline{\textbf{2.600}} & \underline{\textbf{0.093}} & \underline{\textbf{0.969}} & \underline{\textbf{0.992}} &         0.997  \\
    \hline

         \hline
    \end{tabular}
  }
  \caption{\textbf{Quantitative results with new KITTI ground truth, Eigen split and no cropping.} Best results per model in \textbf{bold}, best results across all self-supervised models \underline{underlined}. Experimental results are marked with\expresult, results from original papers with \paperresult. We compare different architectures without TTR, with Ph-TTR and with our SfM-TTR. \textdagger~Results from AdaBins differ from~\cite{bhat2021adabins}, as in this table we do not crop during evaluation. For results using cropping, see Table~\ref{tab:kitti_new_eigen_crop}.} 
  \label{tab:kitti_new_eigen_nocrop}
\end{table*}

\begin{table*}[t]
  \centering
  \footnotesize
  \resizebox{1.0\textwidth}{!}{
   \begin{tabular}{|c|l|c|c|c|c|c|c|c|}\hline
        TTR & Method & Abs Rel~$\downarrow$ & Sq Rel~$\downarrow$ & RMSE~$\downarrow$ & RMSE log~$\downarrow$ & $\delta < 1.25~\uparrow $ & $\delta < 1.25^{2}~\uparrow$ & $\delta < 1.25^{3}~\uparrow$ \\
    \hline\hline

       \xmark  & AdaBins~\cite{bhat2021adabins}\paperresult~\textdagger                               &         0.058  &         0.190  &         2.360  &         0.088  &         0.964  &         0.995  & \textbf{0.999} \\
       \cmark  & \textbf{AdaBins~\cite{bhat2021adabins} + SfM-TTR}~\textdagger            & \textbf{0.054} & \textbf{0.138} & \textbf{1.885} & \textbf{0.078} & \textbf{0.978} & \textbf{0.996} & \textbf{0.999} \\
    \hline
    \hline
       \xmark  & ManyDepth~\cite{watson2021temporal}\expresult                          &         0.059  &         0.297  &         2.960  &         0.097  &         0.954  &         0.991  & \underline{\textbf{0.998}} \\
       \cmark  & ManyDepth~\cite{watson2021temporal} + Ph-TTR\expresult                 & \textbf{0.053} & \textbf{0.252} &         2.774  & \textbf{0.089} &         0.962  & \textbf{0.993} & \underline{\textbf{0.998}} \\
       \cmark  & \textbf{ManyDepth~\cite{watson2021temporal} + SfM-TTR}       &         0.054  & \textbf{0.252} & \textbf{2.510} & \textbf{0.089} & \textbf{0.966} &         0.992  & \underline{\textbf{0.998}} \\
    \hline
       \xmark  & CADepth~\cite{yan2021channel}\expresult                                &         0.073  &         0.359  &         3.287  &         0.112  &         0.941  &         0.990  & \textbf{0.997} \\
       \cmark  & CADepth~\cite{yan2021channel} + Ph-TTR\expresult                       &         0.082  &         0.426  &         3.565  &         0.124  &         0.923  &         0.986  & \textbf{0.997} \\
       \cmark  & \textbf{CADepth~\cite{yan2021channel} + SfM-TTR}             & \textbf{0.060} & \textbf{0.263} & \textbf{2.620} & \textbf{0.096} & \textbf{0.962} & \textbf{0.992} & \textbf{0.997} \\
    \hline
       \xmark  & DIFFNet~\cite{zhou2021diffnet}\expresult                               &         0.066  &         0.318  &         3.078  &         0.103  &         0.953  &         0.992  & \underline{\textbf{0.998}} \\
       \cmark  & DIFFNet~\cite{zhou2021diffnet} + Ph-TTR\expresult                      &         0.053  &         0.252  &         2.778  &         0.090  &         0.965  &         0.993  & \underline{\textbf{0.998}} \\
       \cmark  & \textbf{DIFFNet~\cite{zhou2021diffnet} + SfM-TTR}            & \underline{\textbf{0.052}} & \underline{\textbf{0.229}} & \underline{\textbf{2.444}} & \underline{\textbf{0.085}} & \underline{\textbf{0.973}} & \underline{\textbf{0.994}} & \underline{\textbf{0.998}} \\
    \hline

         \hline
    \end{tabular}
  }
 \caption{\textbf{Quantitative results with new KITTI ground truth, Eigen split and Eigen cropping.} Best results per model in \textbf{bold}, best results across all self-supervised models \underline{underlined}. Experimental results are marked with\expresult, results from papers with \paperresult. \textdagger~Results from AdaBins + SfM-TTR follow the common KITTI Benchmark cropping from the supervised depth learning literature~\cite{bhat2021adabins}, and the AdaBins results without TTR are taken from the original paper.} 
  \label{tab:kitti_new_eigen_crop}
\end{table*}

\begin{table*}[h!]
  \centering
  \footnotesize
  \resizebox{1.0\textwidth}{!}{
    \begin{tabular}{|c|l|c|c|c|c|c|c|c|}\hline
        TTR & Method & Abs Rel~$\downarrow$ & Sq Rel~$\downarrow$ & RMSE~$\downarrow$ & RMSE log~$\downarrow$ & $\delta < 1.25~\uparrow $ & $\delta < 1.25^{2}~\uparrow$ & $\delta < 1.25^{3}~\uparrow$ \\
    \hline\hline

       \xmark  & AdaBins~\cite{bhat2021adabins}\expresult                               & \textbf{0.087} &         0.480  &         3.637  &         0.168  &         0.917  &         0.970  & \textbf{0.985} \\
       \cmark  & \textbf{AdaBins~\cite{bhat2021adabins} + SfM-TTR}            &         0.088  & \textbf{0.454} & \textbf{3.355} & \textbf{0.164} & \textbf{0.927} & \textbf{0.971} & \textbf{0.985} \\
    \hline
    \hline

    \xmark  & Monodepth2 (384x112)~\cite{godard2019digging}\paperresult                          &         \textbf{0.128}  &         \textbf{1.040}  &         5.216      &         0.207  &         0.849  &         \textbf{0.951}  & \textbf{0.978} \\
       \cmark  & Monodepth2 + TTR (from~\cite{luo2020consistent})\paperresult                    &         0.130  &         2.086 &         \textbf{4.876}  & \textbf{0.205} & \textbf{0.878} & 0.946 & 0.970 \\
    \hline

    \xmark  & Monodepth2~\cite{godard2019digging}\paperresult & 0.115  & 0.903  & 4.863 & 0.193 & 0.877  & 0.9590  & 0.981 \\
       \cmark  & Monodepth2 + TTR (from~\cite{tiwari2020pseudo})\paperresult & 0.113 & \textbf{0.793} & 4.655 & 0.188 & 0.874 & 0.960 & \textbf{0.983} \\
    \cmark & \textbf{Monodepth2 + SfM TTR} & \textbf{0.098} & 0.858 & \textbf{4.418} & \textbf{0.177} & \textbf{0.908} & \textbf{0.964} & 0.981\\
    \hline
    
       \xmark  & ManyDepth~\cite{watson2021temporal}\paperresult                          &         0.093  &         0.715  &         4.245  &         0.172  &         0.909  &         0.966  & \textbf{0.983} \\
       \cmark  & ManyDepth~\cite{watson2021temporal} + Ph-TTR\paperresult                 & \underline{\textbf{0.087}} & \textbf{0.696} &         4.183  & \textbf{0.167} & \textbf{0.918} & \textbf{0.968} & \textbf{0.983} \\
       \cmark  & \textbf{ManyDepth~\cite{watson2021temporal} + SfM-TTR}       &         0.090  &         0.718  & \textbf{4.040} &         0.168  &         0.917  &         0.967  & \textbf{0.983} \\
    \hline
       \xmark  & CADepth~\cite{yan2021channel}\paperresult                                &         0.102  &         0.734  &         4.407  &         0.178  &         0.898  & \textbf{0.966} & \underline{\textbf{0.984}} \\
       \cmark  & CADepth~\cite{yan2021channel} + Ph-TTR\expresult                       &         0.110  &         0.802  &         4.648  &         0.187  &         0.878  &         0.962  &         0.983  \\
       \cmark  & \textbf{CADepth~\cite{yan2021channel} + SfM-TTR}             & \textbf{0.095} & \textbf{0.703} & \textbf{4.073} & \textbf{0.173} & \textbf{0.912} & \textbf{0.966} &         0.982  \\
    \hline
       \xmark  & DIFFNet~\cite{zhou2021diffnet}\paperresult                               &         0.097  &         0.722  &         4.345  &         0.174  &         0.907  &         0.967  & \underline{\textbf{0.984}} \\
       \cmark  & DIFFNet~\cite{zhou2021diffnet} + Ph-TTR\expresult                      & \underline{\textbf{0.087}} &         0.667  &         4.138  &         0.167  &         0.920  &         0.968  & \underline{\textbf{0.984}} \\
       \cmark  & \textbf{DIFFNet~\cite{zhou2021diffnet} + SfM-TTR}            & \underline{\textbf{0.087}} & \underline{\textbf{0.660}} & \underline{\textbf{3.948}} & \underline{\textbf{0.165}} & \underline{\textbf{0.925}} & \underline{\textbf{0.969}} & \underline{\textbf{0.984}} \\
    \hline

         \hline
    \end{tabular}
  }
  \caption{\textbf{Quantitative results with Eigen (old) KITTI ground truth, Eigen split and Eigen cropping.} Best results per model in \textbf{bold}, best results across all self-supervised models \underline{underlined}. Experimental results are marked with\expresult, results from original papers with \paperresult. Note how, with this different ground truth, we again outperform the results of the baselines in Tables~\ref{tab:kitti_new_eigen_nocrop} and~\ref{tab:kitti_new_eigen_crop} and we further demonstrate improvement over Monodepth2~\cite{godard2019digging} and the TTR approaches \cite{luo2020consistent,tiwari2020pseudo} that were evaluated after such architecture in the original papers.}
  \label{tab::kitti_old_eigen_crop}
\end{table*}

\section{Limitations}
\label{sec:limitations}

As our current implementation of SfM-TTR depends on COLMAP's output, it is inherently offline and its performance is bounded to the quality of the SfM results. Although we achieve good results in KITTI, a natural scenario and standard benchmark, more challenging setups for SfM (for example, dynamic objects, drastic appearance changes or low-parallax motion) are also problematic for SfM-TTR. Works addressing such SfM challenges~\cite{zhang2022structure} will also be beneficial for our method. Although we could easily replace COLMAP's reconstruction by that of an online real-time visual SLAM pipeline, e.g. ~\cite{campos2021orb}, online and real-time refinement of deep models is not straightforward. We find these aspects relevant for our future work.

Although SfM-TTR excels at medium and large depths, we have noticed a comparable or slightly worse performance than Ph-TTR at very close depths, for which even the adjacent views used in Ph-TTR have sufficient parallax. Observe the metrics in Figure~\ref{fig:error_vs_depth} for depths under 10 meters. This observation suggests a future line of research to combine the best from both Ph-TTR and SfM-TTR.

\vspace{-1mm}
\section{Conclusion}
\label{sec:conclusion}

In this paper we have presented SfM-TTR, an effective test-time refinement for single-view depth networks that preserves the learned priors of supervised and self-supervised models while also leveraging wide-baseline multi-view constraints at inference. The key ingredient is formulating a TTR loss based on sparse SfM depths, which have been estimated from wider baselines than traditional photometric losses, that only consider adjacent frames. We propose a novel RANSAC-based method for scale alignment between SfM and the depth network that accounts for the depth outliers and its heteroscedastic noise. Very importantly, we use a fixed set of hyperparameters for our SfM-TTR for all experiments, without requiring per-architecture or per-sequence tuning. 

Our experiments show that our SfM-TTR improves significantly the depth predictions of different state-of-the-art networks, supervised and self-supervised. We also outperform by a wide margin, in particular at medium and large depths, the common TTR approach that we denote as Ph-TTR, based on the use of photometric losses. These results validate our method as a general TTR approach easy to implement and use after all kinds of networks, current and future ones. Besides, as a more general comment, we believe that the presented contributions provide insights towards a further leverage of SfM in self-supervised depth learning, arising as a promising extension to the widely used photometry-based losses.

\vspace{-1mm}
\section*{Acknowledgments}
\label{sec:acks}

This work was supported by the Spanish Government (PID2021-127685NB-I00 and TED2021-131150B-I00), the Aragón Government (DGA FSE-T45 20R) and the scholarship FPU20/02342.

{\small
\bibliographystyle{ieee_fullname}
\bibliography{egbib}

\begin{thebibliography}{10}\itemsep=-1pt

\bibitem{OpenSfM}
{OpenSfM}.
\newblock \url{https://github.com/mapillary/OpenSfM}.

\bibitem{ali2020real}
Ahmed Ali, Ali Hassan, Afsheen~Rafaqat Ali, Hussam~Ullah Khan, Wajahat Kazmi,
  and Aamer Zaheer.
\newblock Real-time vehicle distance estimation using single view geometry.
\newblock In {\em Proceedings of the IEEE/CVF Winter Conference on Applications
  of Computer Vision}, pages 1111--1120, 2020.

\bibitem{antequera2020mapillary}
Manuel~L{\'o}pez Antequera, Pau Gargallo, Markus Hofinger, Samuel~Rota
  Bul{\`o}, Yubin Kuang, and Peter Kontschieder.
\newblock Mapillary planet-scale depth dataset.
\newblock In {\em European Conference on Computer Vision}, pages 589--604.
  Springer, 2020.

\bibitem{bae2022multi}
Gwangbin Bae, Ignas Budvytis, and Roberto Cipolla.
\newblock Multi-view depth estimation by fusing single-view depth probability
  with multi-view geometry.
\newblock In {\em Proceedings of the IEEE/CVF Conference on Computer Vision and
  Pattern Recognition}, pages 2842--2851, 2022.

\bibitem{barinova2008fast}
Olga Barinova, Vadim Konushin, Anton Yakubenko, KeeChang Lee, Hwasup Lim, and
  Anton Konushin.
\newblock Fast automatic single-view 3-d reconstruction of urban scenes.
\newblock In {\em European Conference on Computer Vision}, pages 100--113.
  Springer, 2008.

\bibitem{bhat2021adabins}
Shariq~Farooq Bhat, Ibraheem Alhashim, and Peter Wonka.
\newblock {AdaBins: Depth Estimation using Adaptive Bins}.
\newblock In {\em Proceedings of the IEEE/CVF Conference on Computer Vision and
  Pattern Recognition}, pages 4009--4018, 2021.

\bibitem{bhat2022localbins}
Shariq~Farooq Bhat, Ibraheem Alhashim, and Peter Wonka.
\newblock {LocalBins: Improving Depth Estimation by Learning Local
  Distributions}.
\newblock In {\em European Conference on Computer Vision}, pages 480--496.
  Springer, 2022.

\bibitem{cadena2016past}
Cesar Cadena, Luca Carlone, Henry Carrillo, Yasir Latif, Davide Scaramuzza,
  Jos{\'e} Neira, Ian Reid, and John~J Leonard.
\newblock Past, present, and future of simultaneous localization and mapping:
  Toward the robust-perception age.
\newblock {\em IEEE Transactions on robotics}, 32(6):1309--1332, 2016.

\bibitem{campos2021orb}
Carlos Campos, Richard Elvira, Juan J~G{\'o}mez Rodr{\'\i}guez, Jos{\'e}~MM
  Montiel, and Juan~D Tard{\'o}s.
\newblock {ORB-SLAM3: An Accurate Open-Source Library for Visual,
  Visual–Inertial, and Multimap SLAM}.
\newblock {\em IEEE Transactions on Robotics}, 37(6):1874--1890, 2021.

\bibitem{casser2019depth}
Vincent Casser, Soeren Pirk, Reza Mahjourian, and Anelia Angelova.
\newblock Depth prediction without the sensors: Leveraging structure for
  unsupervised learning from monocular videos.
\newblock In {\em Proceedings of the AAAI conference on artificial
  intelligence}, volume~33, pages 8001--8008, 2019.

\bibitem{chen2019self}
Yuhua Chen, Cordelia Schmid, and Cristian Sminchisescu.
\newblock Self-supervised learning with geometric constraints in monocular
  video: Connecting flow, depth, and camera.
\newblock In {\em Proceedings of the IEEE/CVF International Conference on
  Computer Vision}, pages 7063--7072, 2019.

\bibitem{eigen2015predicting}
David Eigen and Rob Fergus.
\newblock Predicting depth, surface normals and semantic labels with a common
  multi-scale convolutional architecture.
\newblock In {\em Proceedings of the IEEE international conference on computer
  vision}, pages 2650--2658, 2015.

\bibitem{eigen2014depth}
David Eigen, Christian Puhrsch, and Rob Fergus.
\newblock Depth map prediction from a single image using a multi-scale deep
  network.
\newblock {\em Advances in neural information processing systems}, 27, 2014.

\bibitem{engel2017direct}
Jakob Engel, Vladlen Koltun, and Daniel Cremers.
\newblock Direct sparse odometry.
\newblock {\em IEEE transactions on pattern analysis and machine intelligence},
  40(3):611--625, 2017.

\bibitem{facil2019cam}
Jose~M Facil, Benjamin Ummenhofer, Huizhong Zhou, Luis Montesano, Thomas Brox,
  and Javier Civera.
\newblock {CAM-Convs: Camera-Aware Multi-Scale Convolutions for Single-View
  Depth}.
\newblock In {\em Proceedings of the IEEE/CVF Conference on Computer Vision and
  Pattern Recognition}, pages 11826--11835, 2019.

\bibitem{fischler1981random}
Martin~A Fischler and Robert~C Bolles.
\newblock Random sample consensus: a paradigm for model fitting with
  applications to image analysis and automated cartography.
\newblock {\em Communications of the ACM}, 24(6):381--395, 1981.

\bibitem{fu2018deep}
Huan Fu, Mingming Gong, Chaohui Wang, Kayhan Batmanghelich, and Dacheng Tao.
\newblock Deep ordinal regression network for monocular depth estimation.
\newblock In {\em Proceedings of the IEEE conference on computer vision and
  pattern recognition}, pages 2002--2011, 2018.

\bibitem{geiger2012we}
Andreas Geiger, Philip Lenz, and Raquel Urtasun.
\newblock {Are we ready for autonomous driving? The KITTI vision benchmark
  suite}.
\newblock In {\em 2012 IEEE conference on computer vision and pattern
  recognition}, pages 3354--3361. IEEE, 2012.

\bibitem{godard2017unsupervised}
Cl{\'e}ment Godard, Oisin Mac~Aodha, and Gabriel~J Brostow.
\newblock Unsupervised monocular depth estimation with left-right consistency.
\newblock In {\em Proceedings of the IEEE conference on computer vision and
  pattern recognition}, pages 270--279, 2017.

\bibitem{godard2019digging}
Cl{\'e}ment Godard, Oisin Mac~Aodha, Michael Firman, and Gabriel~J Brostow.
\newblock Digging into self-supervised monocular depth estimation.
\newblock In {\em Proceedings of the IEEE/CVF International Conference on
  Computer Vision}, pages 3828--3838, 2019.

\bibitem{gordon2019depth}
Ariel Gordon, Hanhan Li, Rico Jonschkowski, and Anelia Angelova.
\newblock Depth from videos in the wild: Unsupervised monocular depth learning
  from unknown cameras.
\newblock In {\em Proceedings of the IEEE/CVF International Conference on
  Computer Vision}, pages 8977--8986, 2019.

\bibitem{guizilini2022full}
Vitor Guizilini, Igor Vasiljevic, Rares Ambrus, Greg Shakhnarovich, and Adrien
  Gaidon.
\newblock Full surround monodepth from multiple cameras.
\newblock {\em IEEE Robotics and Automation Letters}, 7(2):5397--5404, 2022.

\bibitem{Hartley2004}
R.~I. Hartley and A. Zisserman.
\newblock {\em Multiple View Geometry in Computer Vision}.
\newblock Cambridge University Press, ISBN: 0521540518, second edition, 2004.

\bibitem{hoiem2005geometric}
Derek Hoiem, Alexei~A Efros, and Martial Hebert.
\newblock Geometric context from a single image.
\newblock In {\em Tenth IEEE International Conference on Computer Vision
  (ICCV'05) Volume 1}, volume~1, pages 654--661. IEEE, 2005.

\bibitem{johnston2020self}
Adrian Johnston and Gustavo Carneiro.
\newblock Self-supervised monocular trained depth estimation using
  self-attention and discrete disparity volume.
\newblock In {\em Proceedings of the ieee/cvf conference on computer vision and
  pattern recognition}, pages 4756--4765, 2020.

\bibitem{karsch2014depth}
Kevin Karsch, Ce Liu, and Sing~Bing Kang.
\newblock Depth transfer: Depth extraction from video using non-parametric
  sampling.
\newblock {\em IEEE transactions on pattern analysis and machine intelligence},
  36(11):2144--2158, 2014.

\bibitem{kingma2014adam}
Diederik~P Kingma and Jimmy Ba.
\newblock Adam: A method for stochastic optimization.
\newblock {\em arXiv preprint arXiv:1412.6980}, 2014.

\bibitem{klodt2018supervising}
Maria Klodt and Andrea Vedaldi.
\newblock Supervising the new with the old: learning sfm from sfm.
\newblock In {\em Proceedings of the European Conference on Computer Vision
  (ECCV)}, pages 698--713, 2018.

\bibitem{kuznietsov2021comoda}
Yevhen Kuznietsov, Marc Proesmans, and Luc Van~Gool.
\newblock Comoda: Continuous monocular depth adaptation using past experiences.
\newblock In {\em Proceedings of the IEEE/CVF Winter Conference on Applications
  of Computer Vision}, pages 2907--2917, 2021.

\bibitem{laina2016deeper}
Iro Laina, Christian Rupprecht, Vasileios Belagiannis, Federico Tombari, and
  Nassir Navab.
\newblock Deeper depth prediction with fully convolutional residual networks.
\newblock In {\em 2016 Fourth international conference on 3D vision (3DV)},
  pages 239--248. IEEE, 2016.

\bibitem{lee2018single}
Jae-Han Lee, Minhyeok Heo, Kyung-Rae Kim, and Chang-Su Kim.
\newblock {Single-Image Depth Estimation Based on Fourier Domain Analysis}.
\newblock In {\em Proceedings of the IEEE Conference on Computer Vision and
  Pattern Recognition}, pages 330--339, 2018.

\bibitem{li2018megadepth}
Zhengqi Li and Noah Snavely.
\newblock {MegaDepth: Learning Single-View Depth Prediction from Internet
  Photos}.
\newblock In {\em Proceedings of the IEEE conference on computer vision and
  pattern recognition}, pages 2041--2050, 2018.

\bibitem{liu2010single}
Beyang Liu, Stephen Gould, and Daphne Koller.
\newblock Single image depth estimation from predicted semantic labels.
\newblock In {\em 2010 IEEE computer society conference on computer vision and
  pattern recognition}, pages 1253--1260. IEEE, 2010.

\bibitem{liu2019neural}
Chao Liu, Jinwei Gu, Kihwan Kim, Srinivasa~G Narasimhan, and Jan Kautz.
\newblock {Neural RGB-D Sensing: Depth and Uncertainty from a Video Camera}.
\newblock In {\em Proceedings of the IEEE/CVF Conference on Computer Vision and
  Pattern Recognition}, pages 10986--10995, 2019.

\bibitem{liu2015learning}
Fayao Liu, Chunhua Shen, Guosheng Lin, and Ian Reid.
\newblock Learning depth from single monocular images using deep convolutional
  neural fields.
\newblock {\em IEEE transactions on pattern analysis and machine intelligence},
  38(10):2024--2039, 2015.

\bibitem{luo2020consistent}
Xuan Luo, Jia-Bin Huang, Richard Szeliski, Kevin Matzen, and Johannes Kopf.
\newblock Consistent video depth estimation.
\newblock {\em ACM Transactions on Graphics (ToG)}, 39(4):71--1, 2020.

\bibitem{mccraith2020monocular}
Robert McCraith, Lukas Neumann, Andrew Zisserman, and Andrea Vedaldi.
\newblock Monocular depth estimation with self-supervised instance adaptation.
\newblock {\em arXiv preprint arXiv:2004.05821}, 2020.

\bibitem{miangoleh2021boosting}
S~Mahdi~H Miangoleh, Sebastian Dille, Long Mai, Sylvain Paris, and Yagiz Aksoy.
\newblock Boosting monocular depth estimation models to high-resolution via
  content-adaptive multi-resolution merging.
\newblock In {\em Proceedings of the IEEE/CVF Conference on Computer Vision and
  Pattern Recognition}, pages 9685--9694, 2021.

\bibitem{moulon2016openmvg}
Pierre Moulon, Pascal Monasse, Romuald Perrot, and Renaud Marlet.
\newblock {OpenMVG: Open Multiple View Geometry}.
\newblock In {\em International Workshop on Reproducible Research in Pattern
  Recognition}, pages 60--74. Springer, 2016.

\bibitem{pan2015inferring}
Jiyan Pan, Martial Hebert, and Takeo Kanade.
\newblock {Inferring 3D Layout of Building Facades From a Single Image}.
\newblock In {\em Proceedings of the IEEE Conference on Computer Vision and
  Pattern Recognition}, pages 2918--2926, 2015.

\bibitem{patil2022p3depth}
Vaishakh Patil, Christos Sakaridis, Alexander Liniger, and Luc Van~Gool.
\newblock {P3Depth: Monocular Depth Estimation with a Piecewise Planarity
  Prior}.
\newblock In {\em Proceedings of the IEEE/CVF Conference on Computer Vision and
  Pattern Recognition}, pages 1610--1621, 2022.

\bibitem{recasens2021endo}
David Recasens, Jos{\'e} Lamarca, Jos{\'e}~M F{\'a}cil, JMM Montiel, and Javier
  Civera.
\newblock {Endo-Depth-and-Motion: Reconstruction and Tracking in Endoscopic
  Videos using Depth Networks and Photometric Constraints}.
\newblock {\em IEEE Robotics and Automation Letters}, 6(4):7225--7232, 2021.

\bibitem{saxena2008make3d}
Ashutosh Saxena, Min Sun, and Andrew~Y Ng.
\newblock {Make3D: Learning 3D Scene Structure from a Single Still Image}.
\newblock {\em IEEE transactions on pattern analysis and machine intelligence},
  31(5):824--840, 2008.

\bibitem{scaramuzza2011visual}
Davide Scaramuzza and Friedrich Fraundorfer.
\newblock Visual odometry [tutorial].
\newblock {\em IEEE robotics \& automation magazine}, 18(4):80--92, 2011.

\bibitem{schonberger2016structure}
Johannes~L Schonberger and Jan-Michael Frahm.
\newblock Structure-from-motion revisited.
\newblock In {\em Proceedings of the IEEE conference on computer vision and
  pattern recognition}, pages 4104--4113, 2016.

\bibitem{shu2020feature}
Chang Shu, Kun Yu, Zhixiang Duan, and Kuiyuan Yang.
\newblock Feature-metric loss for self-supervised learning of depth and
  egomotion.
\newblock In {\em European Conference on Computer Vision}, pages 572--588.
  Springer, 2020.

\bibitem{sturm1999method}
Peter Sturm and Steve Maybank.
\newblock {A Method for Interactive 3D Reconstruction of Piecewise Planar
  Objects from Single Images}.
\newblock In {\em The 10th British machine vision conference (BMVC'99)}, pages
  265--274. The British Machine Vision Association (BMVA), 1999.

\bibitem{tateno2017cnn}
Keisuke Tateno, Federico Tombari, Iro Laina, and Nassir Navab.
\newblock {CNN-SLAM: Real-time dense monocular SLAM with learned depth
  prediction}.
\newblock In {\em Proceedings of the IEEE conference on computer vision and
  pattern recognition}, pages 6243--6252, 2017.

\bibitem{tiwari2020pseudo}
Lokender Tiwari, Pan Ji, Quoc-Huy Tran, Bingbing Zhuang, Saket Anand, and
  Manmohan Chandraker.
\newblock {Pseudo RGB-D for Self-Improving Monocular SLAM and Depth
  Prediction}.
\newblock In {\em European conference on computer vision}, pages 437--455.
  Springer, 2020.

\bibitem{triggs1999bundle}
Bill Triggs, Philip~F McLauchlan, Richard~I Hartley, and Andrew~W Fitzgibbon.
\newblock Bundle adjustment—a modern synthesis.
\newblock In {\em International workshop on vision algorithms}, pages 298--372.
  Springer, 1999.

\bibitem{uhrig2017sparsity}
Jonas Uhrig, Nick Schneider, Lukas Schneider, Uwe Franke, Thomas Brox, and
  Andreas Geiger.
\newblock {Sparsity Invariant CNNs}.
\newblock In {\em 2017 international conference on 3D Vision (3DV)}, pages
  11--20. IEEE, 2017.

\bibitem{watson2021temporal}
Jamie Watson, Oisin Mac~Aodha, Victor Prisacariu, Gabriel Brostow, and Michael
  Firman.
\newblock The temporal opportunist: Self-supervised multi-frame monocular
  depth.
\newblock In {\em Proceedings of the IEEE/CVF Conference on Computer Vision and
  Pattern Recognition}, pages 1164--1174, 2021.

\bibitem{xian2020structure}
Ke Xian, Jianming Zhang, Oliver Wang, Long Mai, Zhe Lin, and Zhiguo Cao.
\newblock Structure-guided ranking loss for single image depth prediction.
\newblock In {\em Proceedings of the IEEE/CVF Conference on Computer Vision and
  Pattern Recognition}, pages 611--620, 2020.

\bibitem{xu2018structured}
Dan Xu, Wei Wang, Hao Tang, Hong Liu, Nicu Sebe, and Elisa Ricci.
\newblock Structured attention guided convolutional neural fields for monocular
  depth estimation.
\newblock In {\em Proceedings of the IEEE conference on computer vision and
  pattern recognition}, pages 3917--3925, 2018.

\bibitem{yan2021channel}
Jiaxing Yan, Hong Zhao, Penghui Bu, and YuSheng Jin.
\newblock Channel-wise attention-based network for self-supervised monocular
  depth estimation.
\newblock In {\em 2021 International Conference on 3D Vision (3DV)}, pages
  464--473. IEEE, 2021.

\bibitem{yang2018deep}
Nan Yang, Rui Wang, Jorg Stuckler, and Daniel Cremers.
\newblock Deep virtual stereo odometry: Leveraging deep depth prediction for
  monocular direct sparse odometry.
\newblock In {\em Proceedings of the European conference on computer vision
  (ECCV)}, pages 817--833, 2018.

\bibitem{yang2018unsupervised}
Zhenheng Yang, Peng Wang, Wei Xu, Liang Zhao, and Ramakant Nevatia.
\newblock Unsupervised learning of geometry from videos with edge-aware
  depth-normal consistency.
\newblock In {\em Proceedings of the AAAI Conference on Artificial
  Intelligence}, volume~32, 2018.

\bibitem{yin2018geonet}
Zhichao Yin and Jianping Shi.
\newblock {GeoNet: Unsupervised Learning of Dense Depth, Optical Flow and
  Camera Pose}.
\newblock In {\em Proceedings of the IEEE conference on computer vision and
  pattern recognition}, pages 1983--1992, 2018.

\bibitem{yuan2022neural}
Weihao Yuan, Xiaodong Gu, Zuozhuo Dai, Siyu Zhu, and Ping Tan.
\newblock Neural window fully-connected crfs for monocular depth estimation.
\newblock In {\em Proceedings of the IEEE/CVF Conference on Computer Vision and
  Pattern Recognition}, pages 3916--3925, 2022.

\bibitem{zaheer2018single}
Aamer Zaheer, Maheen Rashid, Muhammad~Ahmed Riaz, and Sohaib Khan.
\newblock Single-view reconstruction using orthogonal line-pairs.
\newblock {\em Computer Vision and Image Understanding}, 172:107--123, 2018.

\bibitem{zhan2018unsupervised}
Huangying Zhan, Ravi Garg, Chamara~Saroj Weerasekera, Kejie Li, Harsh Agarwal,
  and Ian Reid.
\newblock Unsupervised learning of monocular depth estimation and visual
  odometry with deep feature reconstruction.
\newblock In {\em Proceedings of the IEEE conference on computer vision and
  pattern recognition}, pages 340--349, 2018.

\bibitem{zhang2009consistent}
Guofeng Zhang, Jiaya Jia, Tien-Tsin Wong, and Hujun Bao.
\newblock Consistent depth maps recovery from a video sequence.
\newblock {\em IEEE Transactions on pattern analysis and machine intelligence},
  31(6):974--988, 2009.

\bibitem{zhang2022structure}
Zhoutong Zhang, Forrester Cole, Zhengqi Li, Michael Rubinstein, Noah Snavely,
  and William~T Freeman.
\newblock Structure and motion from casual videos.
\newblock In {\em European Conference on Computer Vision}, pages 20--37.
  Springer, 2022.

\bibitem{zhou2021diffnet}
Hang Zhou, David Greenwood, and Sarah Taylor.
\newblock Self-supervised monocular depth estimation with internal feature
  fusion.
\newblock In {\em British Machine Vision Conference (BMVC)}, 2021.

\bibitem{zhou2017unsupervised}
Tinghui Zhou, Matthew Brown, Noah Snavely, and David~G Lowe.
\newblock Unsupervised learning of depth and ego-motion from video.
\newblock In {\em Proceedings of the IEEE conference on computer vision and
  pattern recognition}, pages 1851--1858, 2017.

\bibitem{zhou2022self}
Zhengming Zhou and Qiulei Dong.
\newblock Self-distilled feature aggregation for self-supervised monocular
  depth estimation.
\newblock In {\em European Conference on Computer Vision}, pages 709--726.
  Springer, 2022.

\bibitem{zhu2022autonomous}
Pengli Zhu, Siyuan Liu, Tao Jiang, Yancheng Liu, Xuzhou Zhuang, and Zhenrui
  Zhang.
\newblock Autonomous reinforcement control of visual underwater vehicles:
  Real-time experiments using computer vision.
\newblock {\em IEEE Transactions on Vehicular Technology}, 71(8):8237--8250,
  2022.

\end{thebibliography}
}

\balance

\end{document}